\documentclass[preprint,authoryear, english,fleqn]{elsarticle}
\usepackage{enumerate}
\usepackage{amsmath}
\usepackage{multirow}
\usepackage{threeparttable}
\usepackage[T1]{fontenc}
\usepackage{array}
\usepackage{longtable}
\usepackage{graphicx}
\usepackage{esint}
\usepackage{float}
\usepackage{caption}
\usepackage{subcaption}
\usepackage{amssymb}
\usepackage{amsthm}

\usepackage{array,booktabs}
\usepackage{verbatim}

\makeatletter
\AtBeginDocument{}
\providecommand{\\}{\\}
\@ifundefined{date}{}{\date{}}
\usepackage{epstopdf}
\DeclareGraphicsExtensions{.pdf,.jpeg,.png}
\usepackage[paperheight=11.69in,paperwidth=8.27in,left=1.2in,right=1.2in,top=1.2in,bottom=1.2in,headheight=1.2in]{geometry}
\usepackage[hidelinks]{hyperref}

\def\equationautorefname~#1\null{Eq.~(#1)\null}
\usepackage{cleveref}
\crefrangeformat{equation}{Eqs. #3(#1)#4--#5(#2)#6}
\creflabelformat{equation}{#2\textup{#1}#3}
\usepackage{longtable}
\makeatother
\usepackage{babel}
\usepackage{graphicx}
\graphicspath{ {./figure/} }
\allowdisplaybreaks[4]
\usepackage[textfont=normalsize]{subcaption}
\usepackage[font=normalsize]{caption}
\usepackage{enumitem}
\setlist[enumerate,1]{label=\textup{(\roman{*})}, parsep = 0pt, itemindent=0.1cm, itemsep=0pt}
\usepackage{adjustbox}
\usepackage{tabularx}
\usepackage{tabu}

\makeatletter
\def\ps@pprintTitle{%
   \let\@oddhead\@empty
   \let\@evenhead\@empty
   \def\@oddfoot{\reset@font\hfil\thepage\hfil}
   \let\@evenfoot\@oddfoot
}
\makeatother
\def\eqref#1{{(\ref{#1})}}

\usepackage[mathlines,running]{lineno}
\expandafter\let\expandafter\oldequation\csname equation\endcsname
\expandafter\let\expandafter\oldgather\csname gather\endcsname
\expandafter\let\expandafter\endoldequationstar\csname endequation*\endcsname
\expandafter\def\csname equation\endcsname{%
  \ifLineNumbers%
  \expandafter\linenomath%
  \fi%
  \oldequation%
}
\expandafter\def\csname gather\endcsname{%
  \ifLineNumbers%
  \expandafter\linenomath%
  \fi%
  \oldgather%
}
\expandafter\def\csname endequation*\endcsname{%
  \endoldequationstar%
  \ifLineNumbers%
  \def\maybeendlinenomath{\expandafter\endlinenomath}%
  \else
  \def\maybeendlinenomath{}%
  \fi%
  \expandafter\maybeendlinenomath%
}

\begin{document}
\begin{frontmatter}{}
\title{A Bibliometric Analysis and Review on Reinforcement Learning for Transportation Applications}



\author[a]{Can Li}
\address[a]{School of Computer Science and Engineering, University of New South Wales, Sydney, NSW 2052, Australia}

\author[b]{Lei Bai}
\address[b]{School of Electrical and Information Engineering, University of Sydney, Sydney, NSW 2008, Australia}

\author[a]{Lina Yao}

\author[c]{S. Travis Waller}
\address[c]{Lighthouse Professorship ``Transport Modelling and Simulation”, Faculty of Transport and Traffic Sciences, Technische Universit{\"a}t Dresden, Germany}

\author[d]{Wei Liu\corref{correspond_author}}\cortext[correspond_author]{Corresponding author}
\ead{wei.w.liu@polyu.edu.hk}
\address[d]{Department of Aeronautical and Aviation Engineering, The Hong Kong Polytechnic University, Hong Kong, China}

\begin{abstract}
Transportation is the backbone of the economy and urban development. Improving the efficiency, sustainability, resilience, and intelligence of transportation systems is critical and also challenging. The constantly changing traffic conditions, the uncertain influence of external factors (e.g., weather, accidents), and the interactions among multiple travel modes and multi-type flows result in the dynamic and stochastic natures of transportation systems. The planning, operation, and control of transportation systems require flexible and adaptable strategies in order to deal with uncertainty, non-linearity, variability, and high complexity. In this context, Reinforcement Learning (RL) that enables autonomous decision-makers to interact with the complex environment, learn from the experiences, and select optimal actions has been rapidly emerging as one of the most useful approaches for smart transportation. This paper conducts a bibliometric analysis to identify the development of RL-based methods for transportation applications, typical journals/conferences, and leading topics in the field of intelligent transportation in recent ten years. Then, this paper presents a comprehensive literature review on applications of RL in transportation by categorizing different methods with respect to the specific application domains. The potential future research directions of RL applications and developments are also discussed.
\end{abstract}


\end{frontmatter}{}

\section{Introduction} \label{sec: introduction}

The travel demand is increasing along with the growth of social and economic activities, which results in great challenges in terms of crowding, congestion, emission, energy, and safety. Meanwhile, the massive amount of multi-source data has been continuously and/or automatically collected. In this context, artificial intelligence (AI) methods have been proposed to take advantage of the growing data availability in order to address challenges faced by transportation systems and travelers and thus improve system safety, sustainability, resilience, and efficiency.

Reinforcement Learning (RL) is an essential branch of AI-based methods, which is an experience-driven autonomous learning strategy and is often formulated based on Markov Decision Processes (MDPs). RL can be regarded as a process where the agent learns optimal behaviors/decisions by trial-and-error interactions with the environment \citep{kaelbling1996reinforcement}. It is more practical than supervised learning methods in many occasions, which does not necessarily require prior experiences or sufficient historical data to train the agent \citep{ye2019automated}. Some RL-based models are quite well scalable to high-dimensional systems \citep{desjardins2011cooperative}, making them adaptable to complex problems based on simple instances. Moreover, RL-based approaches that do not need re-optimization when changes occur in the environment may save computation efforts and increase practicality \citep{zhou2019development}. Also, the RL-based strategy is capable of capturing the long-term effect of current actions and achieving greater efficiency and profits \citep{pan2019deep}. The aforementioned advantages of Reinforcement Learning indeed attract substantial research efforts that adopt and develop RL-based models for decision-making, especially in game playing. For example, the DeepMind team first applies RL on Atari 2600 games to learn optimal policies and achieves better performance than human players \citep{mnih2015human}. AlphaGo \citep{silver2016mastering} verifies the superiority of Reinforcement Learning with an extremely high winning ratio. 

In line with the success of Reinforcement Learning in the field of game playing, many studies have developed and/or applied RL strategies in the transportation sector. The experimental results evaluating on real-world datasets or synthetic datasets demonstrate the effectiveness of Reinforcement Learning in learning and managing transportation systems, improving accuracy and efficiency, and reducing resource consumption. There are several existing reviews on RL studies in the transportation domain. In particular, \citet{mannion2016experimental} and \citet{yau2017survey} focus on traffic signal control with RL; \citet{aradi2020survey} and \citet{kiran2021deeprl} focus on deep RL models for autonomous driving. Three additional review studies \citep{abdulhai2003survey, haydari2020deeprl, farazi2021deep} have covered more transportation applications with Reinforcement Learning. \citet{abdulhai2003survey} was published in 2003, which does not cover the substantial development of RL methods in transportation in recent years. \citet{farazi2021deep} mainly focuses on deep RL methods for applications in transportation (e.g., autonomous driving and traffic signal control). However, non-deep RL models have not been examined. \citet{haydari2020deeprl} has discussed both deep RL and non-deep RL methods and covers a wide range of RL applications in transportation (including traffic signal control, energy management for electric vehicle, road control, and autonomous driving). However, the importance of fairness in developing RL methods for transportation applications is ignored in previous works. Moreover, none has provided a bibliometric analysis of RL methods for transportation applications. Differently, this study takes advantage of the bibliometric analysis to provide a systematic review on applications of both deep RL and non-deep RL methods in transportation, and provide more comprehensive coverage of applications than related existing reviews (e.g., including RL applications in taxi and bus systems that have not been covered by \citet{haydari2020deeprl}). Besides, this paper summarizes several aspects that require substantial efforts in terms of developing RL methods for real-world transportation
applications, i.e., scalability, practicality, transferability, and fairness.

\begin{figure}[htbp]
    \centering
    \includegraphics[width=1.1\linewidth]{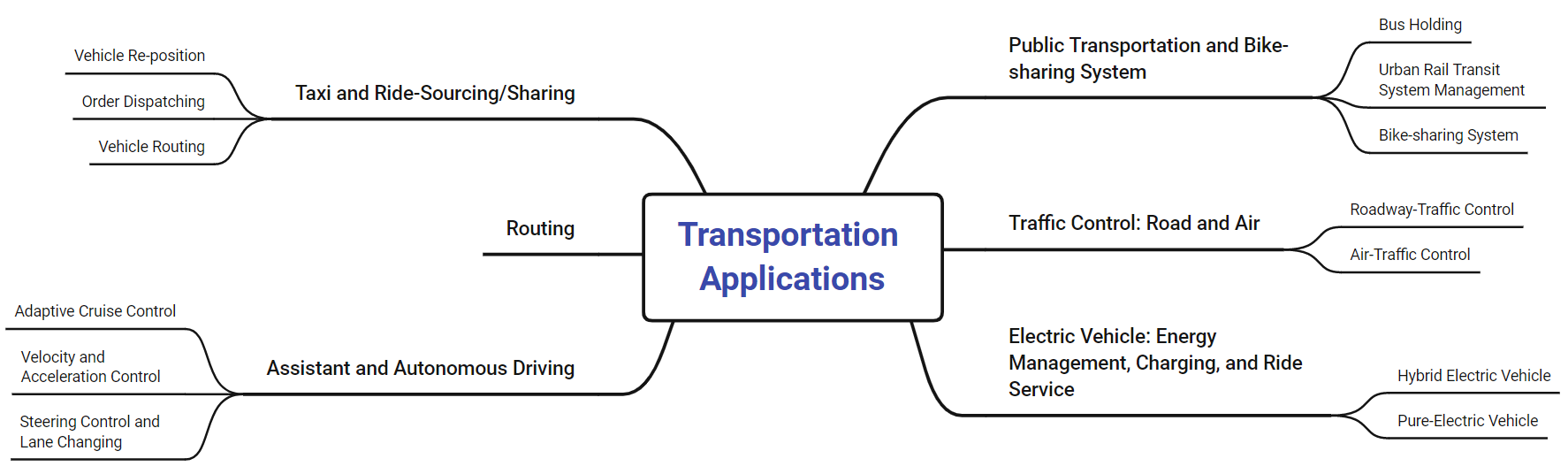}
    \caption{Classification of RL Applications in Transportation}
    \label{fig:framework}
\end{figure}

Specifically, this study provides a summary on applications of RL to address relevant transportation issues and takes advantage of the bibliometric analysis approach to construct network connections of the journals/conferences and keywords to identify the influential journals/conferences and areas of concern. Several future directions of RL studies in transportation are also discussed. The major transportation topics that involve RL methods discussed in this study include traffic control, taxi and ride-sourcing/sharing, assistant and autonomous driving, routing, public transportation and bike-sharing system, and electric vehicles. The detailed classification of topics is shown in Fig.~\ref{fig:framework}. This review collects over a hundred of related papers mostly published in the last ten years in major journals in the transportation domain (e.g., Transportation Research Part B, Part C, IEEE Transactions on Intelligent Transportation Systems, IET Intelligent Transport Systems) and major related conferences in the computer science domain (e.g., AAAI, KDD, WWW, CIKM), which will be discussed in Section~\ref{sec:ana}. To summarize, this paper provides a reference point to researchers for interdisciplinary Reinforcement Learning research in transportation and computer science.

The rest of this paper is structured as follows. Section~\ref{sec:preliminary} introduces the basic formulations of Reinforcement Learning and Section~\ref{sec:ana} conducts the bibliometric study. The review of the six topic categories for transportation applications with RL are presented in Section~\ref{sec:traffic_control} $-$ Section~\ref{sec:ee}, respectively. Future directions of RL in transportation and the conclusion of this paper are discussed in Section~\ref{sec:challenge}.

\section{Preliminary} \label{sec:preliminary}

This section presents the basic formulation of Markov Decision Process (MDP) and Reinforcement Learning, and the usage of data for RL algorithms. Three main algorithms in RL are also summarized, i.e., Value-based RL, Policy-based RL, and Actor-Critic-based RL.

\subsection{Markov Decision Process}

Reinforcement Learning is formulated based on Markov Decision Process (MDP), a framework applied in stochastic control theory \citep{sutton2018reinforcement}. MDP consists of five elements, $<\mathcal{S, A, P, R, \gamma}>$ where $\mathcal{S}$ represents the set of states, $\mathcal{A}$ denotes the set of actions, $\mathcal{P}$ is the probabilistic transition function, $\mathcal{R}$ is the reward function, and $\mathcal{\gamma}\in[0,1]$ denotes the discount factor. At time step $t$, under a state $s_{t}\in\mathcal{S}$, the agent performs an action $a_{t}\in\mathcal{A}$ and then receives an immediate reward $r_{t}(s_{t},a_{t})\in\mathcal{R}$ from the environment. The environment state will change to $s_{t+1}\in\mathcal{S}$ based on the transition probability $\mathcal{P}(s_{t+1}|s_{t},a_{t})$. The goal of the agent is to find a optimal policy $\pi^{*}$ for maximizing the cumulative reward with discount factor where $\mathcal{G} = \sum_{t=1}^{T}\mathcal{\gamma}^{t}r_{t}$, $\pi^{*} = \text{argmax}_{\pi} \mathbb{E}[\mathcal{G}|\pi]$, and $\mathbb{E}$ represents the expected value.

\subsection{Reinforcement Learning Algorithms}

Based on the policy $\pi$, in order to evaluate the current state and according action in RL, the state-value function $V^{\pi}(s)$ and state-action value function $Q(s, a)$ are introduced below:
\begin{align}
    V^{\pi}(s) &= \mathbb{E}[\mathcal{G} | s] \\
    Q^{\pi}(s, a) &= \mathbb{E}[\mathcal{G} | s, a] \\
    V^{\pi}(s) &= \sum_{a} \pi(a|s)Q(s, a) \\
    Q^{\pi}(s, a) &= \sum_{s^{\prime}}\mathcal{P}(s^{\prime}|s, a)(r(s,a) + V(s^{\prime}))
\end{align}
The optimal policy can be obtained by letting $\pi(s) = \text{argmax}_{a} Q(s, a)$ and the state-value function is $V^{\pi}(s) = \max_{a}Q^{\pi}(s,a)$. Then, Bellman Expectation Equation \citep{bellman1952theory} can be used to solve the value function:
\begin{equation}
    V^{\pi}(s) = \sum_{a}\pi(a|s)\sum_{s^{\prime}, r}p(s^{\prime},r|s,a)[r + \gamma V^{\pi}(s^{\prime})]
\end{equation}

Policy iteration and value iteration are utilized to solve $\pi$ and $V^{\pi}(s)$. Thus, according to the mechanism of iteration, RL can be divided into three categories, i.e., the Value-based RL, Policy-based RL, and Actor-Critic-based RL. The basic definitions of these three policies are introduced below.

\subsubsection{Value-based Reinforcement Learning}

In the value iteration approach, the value function is updated following the Bellman Optimal Equation \citep{bellman1952theory}:
\begin{equation}
    V_{k+1}(s) = \max_{a} \mathbb{E}[r_{t+1} + \gamma V_{k}(S_{t+1})|(S_{t} = s, A_{t} = a)]
\label{Eq:bellman_optimal}
\end{equation}
Two classic approaches have been used to estimate $V^{\pi}(s)$, Monte-Carlo-based approach (MC) and Temporal-Difference-based approach (TD). In MC, based on current state $s(t)$, the agent starts to interact with the environment until reaching a termination condition. Then, the cumulative reward $\mathcal{G}_{t}$ can be calculated based on given rules. The aim is to drive $V_{t}(s)$ close to $\mathcal{G}_{t}$, which leads to the update policy as follow:
\begin{equation}
    V_{t}(s) \gets V_{t}(s) + \alpha(\mathcal{G}_{t} - V_{t}(s))
\end{equation}
where $\alpha$ is the learning rate. Since the reward obtained by MC is estimated at the end of the episode, there can be large variances in the cumulative reward. On the contrary, TD only simulates one step in the episode and the update policy is as follows:
\begin{equation}
    V_{t}(s) \gets V_{t}(s) + \alpha(\mathcal{R}_{t} + \gamma V_{t}(s+1) - V_{t}(s))
\label{Eq:TD}
\end{equation}
which yields smaller variance but can be less accurate due to a lack of an overview of the whole episode.

Typical TD-based strategies are Q-learning \citep{watkins1992q} and State-Action-Reward-State-Action (Sarsa) algorithm \citep{sutton1996generalization} which replace $V^{\pi}(s)$ with $Q(s,a)$ following Eq.~\eqref{Eq:TD}. The update policy of Q-learning is represented as:
\begin{equation}
    Q^{\pi}(s_{t},a_{t}) \gets Q^{\pi}(s_{t},a_{t}) + \alpha (r_{t} + \gamma max_{a_{t+1}}Q^{\pi}(s_{t+1}, a_{t+1}) - Q^{\pi}(s_{t},a_{t}))
\end{equation}
And the update policy of Sarsa can be shown as:
\begin{equation}
    Q^{\pi}(s_{t},a_{t}) \gets Q^{\pi}(s_{t},a_{t}) + \alpha (r_{t} + \gamma Q^{\pi}(s_{t+1}, a_{t+1}) - Q^{\pi}(s_{t},a_{t}))
\end{equation}
Both Q-learning and Sarsa have two policies, a behavior policy to interact with the environment and sample potential actions from the learning data with randomness and a target policy to improve the performance with the help of sampling data and thus obtain the optimal policy. Furthermore, according to the data usage when updating value functions, RL can be divided into on-policy and off-policy methods. On-policy methods update the policy that is used to make decisions, while off-policy methods update a policy different from that used to generate the data \citep{sutton1998introduction}. Sarsa is an on-policy strategy (i.e., the target policy is the same as the behavior policy), while Q-learning is an off-policy method (i.e., the target policy is to suppose the selecting action with the largest reward to update the value function).

In some applications, a large number of states and actions can hardly be captured in Q-learning. Thus, deep models are used to approximate the value function. \citet{mnih2015human} proposes Deep Q-Network (DQN) for optimal policy finding. Given a Q-function $Q$ and a target Q-function $\hat{Q}$ initializing by $\hat{Q}=Q$, an experience replay buffer is utilized to store the transition $(s_{t}, a_{t}, r_{t}, s_{t+1})$ in each time step where $a_{t}$ is obtained by $Q$. When enough sample data is obtained, a mini-batch of samples is chosen randomly to get the target value by $\hat{Q}$:
\begin{equation}
    y = r_{i} + \gamma \max_{a}\hat{Q}(s_{i+1}, a)
\end{equation}
Then, the parameters of $Q$ are updated by driving $Q(s_{i}, a_{i})$ close to $y$ with the gradient descent method. The target network $\hat{Q}$ will be reset after $C$ steps by $\hat{Q}=Q$. 

Further DQN-based methods such as Double-DQN \citep{van2016deep} and Dueling-DQN \citep{wang2016dueling} are developed for more robust and faster policy learning. In detail, to reduce the overestimations caused by DQN (i.e., the estimated value is larger than the true value), Double-DQN implements the choice and the evaluation of actions with different value functions, $Q^{A}(s,a)$ and $Q^{B}(s,a)$. Thus, the updated function can be represented as:
\begin{equation}
\begin{aligned}
Q^{B}(s_{t},a_{t}) \gets Q^{A}(s_{t},a_{t}) + \alpha (r_{t} + \gamma max_{a_{t+1}}Q^{B}(s_{t+1}, argmax_{a}Q^{A}(s_{t+1},a_{t})) - Q^{A}(s_{t},a_{t})) \\
Q^{B}(s_{t},a_{t}) \gets Q^{B}(s_{t},a_{t}) + \alpha (r_{t} + \gamma max_{a_{t+1}}Q^{A}(s_{t+1}, argmax_{a}Q^{B}(s_{t+1},a_{t})) - Q^{B}(s_{t},a_{t}))
\end{aligned}
\end{equation}
Moreover, Dueling-DQN divides the action value function into state value function and advantage function, i.e., $Q^{\pi}(s_{t},a_{t}) = V^{\pi}s_{t} + A^{\pi}(s_{t},a_{t})$ where $A^{\pi}(s_{t},a_{t})$ denotes as the advantage function for the strategy evaluation. The advantage function is larger than zero means that the action is better than the average action, otherwise the current action is worse than the average action.

\subsubsection{Policy-based Reinforcement Learning}

Value-based RL models are often effective in discrete control tasks but are hard to be adapted into continuous action space \citep{arulkumaran2017deep}. Policy-based RL can help solve such issues and is more applicable in high dimensional action spaces. 

\citet{sutton2000policy} introduces the Policy Gradient method where the policy is written as $\pi_{\theta}(s,a)$ with the parameter vector $\theta$. The objective is to maximize the average reward $\rho(\pi)$ which can be obtained by the policy gradient:
\begin{equation}
    \frac{\partial \rho}{\partial \theta} = \sum_{s} d^{\pi}(s) \sum_{a}\frac{\partial \pi(s, a)}{\partial \theta} Q^{\pi}(s,a)
\end{equation}
where $d^{\pi}(s,a) = \sum_{t=0}^{\infty} \gamma^{t}Pr(s_{t}=s|s_{0},\pi)$ and $Q^{\pi}(s,a) = \mathbb{E}[\sum_{k=1}^{\infty}\gamma^{k-1}r_{t+k}|s_{t}=s, a_{t}=a, \pi]$. Then, the Policy Gradient with Function Approximation can be written as:
\begin{equation}
    \frac{\partial \rho}{\partial \theta} = \sum_{s} d^{\pi}(s) \sum_{a}\frac{\partial \pi(s, a)}{\partial \theta} f_{w}(s,a)
\end{equation}
where $\frac{\partial f_{w}(s,a)}{\partial w} = \frac{\partial \pi(s, a)}{\partial \theta} \frac{1}{\pi(s,a)}$.

Proximal Policy Optimization (PPO) \citep{schulman2017proximal} is an improved and widely adopted on-policy algorithm which solves the problem that the choice of step size in the Policy Gradient algorithm is not straightforward.

\subsubsection{Actor-Critic-based Reinforcement Learning}

Actor-Critic-based (AC-based) RL \citep{sutton2000policy} takes advantage of both value-based function and policy-based function. The actor network interacts with the environment and generates actions. The critic network uses the value function to evaluate the performance of the actor and guide the actor's actions in the next time step.

Some widely-used algorithms in AC-based RL are Deterministic Policy Gradient (DPG) \citep{silver2014deterministic}, Deep Deterministic Policy Gradient (DDPG) \citep{timothy2016continuous}, Advantage Actor-Critic (A2C), and Asynchronous Advantage Actor-Critic (A3C) \citep{babaeizadeh2017reinforcement}. DPG and DDPG are off-policy methods that can be easier to train in high dimensional action space, and DDPG is based on deep learning. A2C and A3C are on-policy algorithms where A2C adopts synchronous control method, and A3C adopts asynchronous control method for actor network updating. A3C is often adopted in transportation problems for policy-making, which is introduced in detail as an example to illustrate the mechanism of synchronous methods. A3C takes advantage of the Actor-Critic framework and introduces the synchronous method to improve the performance and efficiency. Multiple threads are utilized in A3C to collect data in parallel, i.e., each thread is an independent agent to explore an independent environment, and each agent can use different strategies to sample data. Sampling data independently is able to obtain unrelated samples and increase sampling speed.

\subsection{Data Usage}

Both synthetic data and real-world data are largely used in studies for transportation applications with RL. On the one hand, it is easier and more feasible to obtain synthetic data. A large number of scenarios/samples with different characteristics can be constructed to evaluate the proposed methods. However, some uncertainties, disruptions, and accidents occurring in practice are hard to be measured or simulated, which leaves a certain and unknown gap with actual environments. On the other hand, the real-world data can reflect the actual situations more accurately, which means the proposed method can be put into practice for the scenario corresponding to the collected data. It is harder to obtain complete and diverse real-world data due to several reasons, e.g., the confidentiality of various sources and the lack of information. Also, a real-world dataset only represents the characteristics of a specific target, which has limited scenarios/samples to evaluate the generality of proposed models.

Although the applications and corresponding data are diverse, the type of data can be divided into three categories, i.e., road network relevant data, traffic flow relevant data, and vehicle operation relevant data. And the usage of the data in the algorithms is in a similar way. Specifically, road networks are regarded as directed graphs with nodes and edges (i.e., nodes denote intersections while edges represent roads). Some other road related characteristics (e.g., speed limit, the number of lanes/tracks, and distributions of stations) are also concluded to construct the stationary environment of RL. The traffic flow relevant data (e.g., traffic speed and demand) and vehicle operation relevant data (e.g., fuel/electricity consumption and lane changing) are used as the time-varying input of RL models to constitute the dynamic environment of RL. The agents learn and analyze the information of both stationary and dynamic environments for decision-making based on many different RL-based optimization strategies.

\begin{table}[htbp]
\caption{Numbers of Related Publications in Major Journals/Conferences (as of December 31, 2021)}
\setlength{\tabcolsep}{.4mm}{
\begin{tabular}{c|c|c}
\hline
\textbf{Attribute} & \textbf{Name} & \textbf{\begin{tabular}[c]{@{}c@{}}Number of \\ Related Papers\end{tabular}} \\ \hline
Conference & \begin{tabular}[c]{@{}c@{}} IEEE International Conference on Intelligent \\ Transportation Systems (ITSC) \end{tabular} & 124 \\ \hline
Journal & IEEE Transactions on Vehicular Technology (T-VT) & 119 \\ \hline
Journal & Transportation Research Part C: Emerging Technologies (TR-C) & 28 \\ \hline
Journal & \begin{tabular}[c]{@{}c@{}} IEEE Transactions on Intelligent \\ Transportation Systems (T-ITS) \end{tabular} & 27 \\ \hline
Journal & IET Intelligent Transport Systems & 14 \\ \hline
Conference & Association for the Advancement of Artificial Intelligence (AAAI) & 11 \\ \hline
Journal & \begin{tabular}[c]{@{}c@{}} Transportation Research Record: \\ Journal of the 
Transportation Research Board \end{tabular} & 9 \\ \hline
Conference & \begin{tabular}[c]{@{}c@{}} Special Interest Group on Knowledge Discovery \\ and Data Mining (SIGKDD) \end{tabular} & 7 \\ \hline
Journal & IEEE Transactions on Transportation Electrification & 6 \\ \hline
Conference & International Joint Conference on Artificial Intelligence (IJCAI) & 6 \\ \hline
Conference & Conference on Information and Knowledge Management (CIKM) & 4 \\ \hline
Conference & World Wide Web Conference (WWW) & 4 \\ \hline
Journal & Transportation Research Part B: Methodological (TR-B) & 3 \\ \hline
Journal & \begin{tabular}[c]{@{}c@{}} Transportation Research Part E: \\ Logistics and Transportation Review (TR-E) \end{tabular} & 3 \\ \hline
Conference & International Conference on Data Mining (ICDM) & 3 \\ \hline
Journal & Transportmetrica B: Transport Dynamics & 1 \\ \hline
Journal & Transportation & 1 \\ \hline
Journal & Transportation Science & 1 \\ \hline
Journal & \begin{tabular}[c]{@{}c@{}} Transportation Research Part F: \\ Traffic Psychology and Behaviour (TR-F) \end{tabular} & 1 \\ \hline
Journal & Journal of Transportation Engineering Part A: Systems & 1 \\ \hline
Journal & Research in Transportation Economics & 1 \\ \hline
Journal & Transport Reviews & 0 \\ \hline
Journal & Transportation Research Part A: Policy and Practice (TR-A) & 0 \\ \hline
Journal & \begin{tabular}[c]{@{}c@{}} Transportation Research Part D: \\ Transport and Environment (TR-D) \end{tabular} & 0 \\ \hline
Journal & Journal of Transport Geography & 0 \\ \hline
Journal & Journal of Air Transport Management & 0 \\ \hline
Journal & Travel Behaviour and Society & 0 \\ \hline
Journal & Transportmetrica A: Transport Science & 0 \\ \hline
Journal & Transport Policy & 0 \\ \hline
Journal & International Journal of Sustainable Transportation & 0 \\ \hline
Journal & Maritime Policy $\&$ Management & 0 \\ \hline
Journal & Journal of Transportation Engineering, Part B: Pavements & 0 \\ \hline
\end{tabular}}
\label{table:number}
\end{table}

\section{Bibliometric Analysis} \label{sec:ana}

This section provides a bibliometric analysis of studies for RL-based transportation applications. The distribution of published papers in journals/conferences and the characteristics of research fields or topics are explored. The VOSviewer software \footnote{https://www.vosviewer.com/} is used to measure the influences and relations of publications and keywords.

The selected journals and conferences covering January 2010 to December 2021 are summarized in Table~\ref{table:number} according to the number of published related papers. The list of journals and conferences is based on the following. The selected transportation-related journals are ranked as Q1, Q2, and Q3 by Scimago Journal $\&$ Country Rank in 2021.\footnote{https://www.scimagojr.com/journalrank.php} The selected conferences in the field of artificial intelligence and data mining are with the highest CORE ranking (CORE A+) in recent years.\footnote{http://cic.tju.edu.cn/faculty/zhileiliu/doc/COREComputerScienceConferenceRankings.html}. International Conference on Intelligent Transportation Systems (ITSC) is also included due to its high relevance and wide audience. It can be seen that ITSC covers a substantial number of RL-based transportation applications studies (nearly one-third of the total number of publications, i.e., about 33.16\%), which indicates that Reinforcement Learning has attracted substantial attentions for achieving intelligent traffic control and management. Following ITSC are T-VT, TR-C, and T-ITS with 119 (31.82\%), 28 (7.49\%), and 27 (7.22\%) papers, respectively, which indicates the fusion and interaction of traditional transportation applications and popular machine learning strategies over the recent decade. Several transportation journals involve a relatively small number of papers regarding applications of RL (e.g., TR-A, TR-D, and Transportmetrica A), indicating that there are significant research gaps here for developing advanced RL in diverse aspects of transportation.

\begin{figure}[htbp]
    \centering
    \includegraphics[width=.75\linewidth]{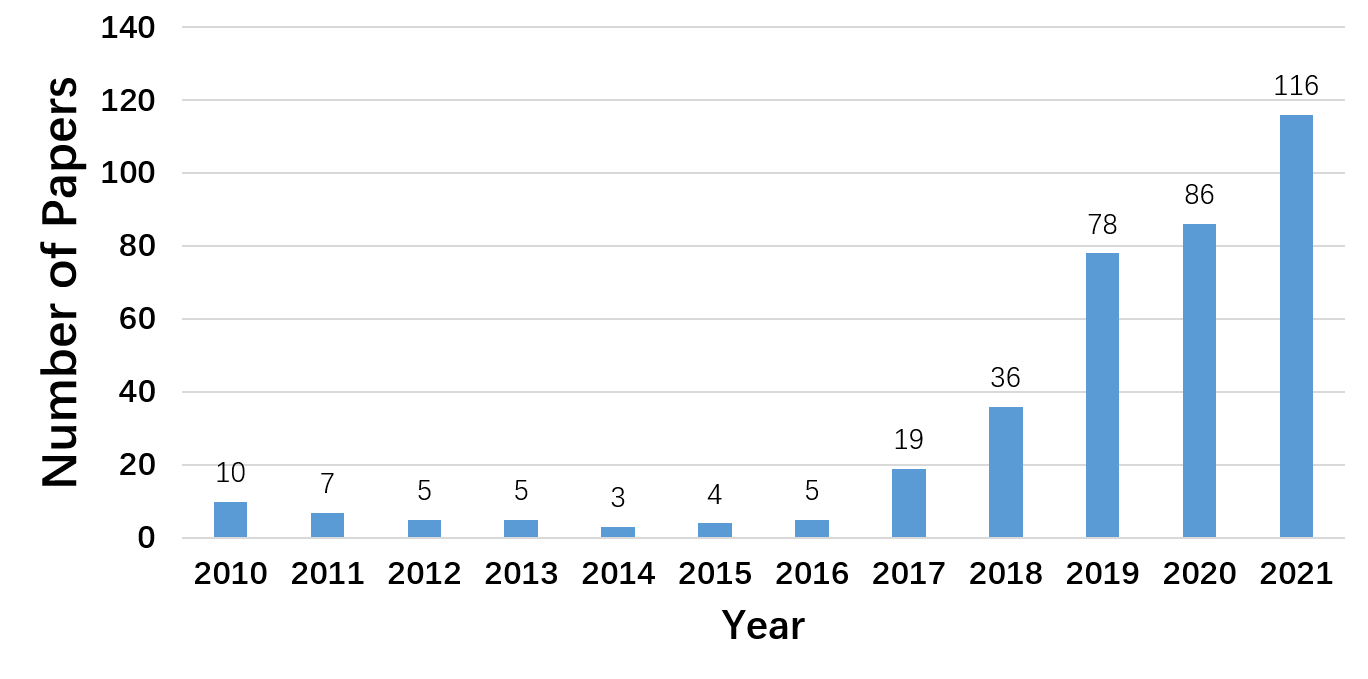}
    \caption{the Number of Published Related Papers per Year (Jan. 2010 - Dec. 2021)}
    \label{fig:year}
\end{figure}

In addition, the numbers of the published papers in the aforementioned journals and conferences from January 2010 to December 2021 are shown in Fig.~\ref{fig:year}. Before 2017, only a few studies per year focused on Reinforcement Learning to solve transportation problems, with only 39 articles published in total in the selected journals and conferences. And the number of published related papers from 2011 to 2016 is between three and seven (around five), which is regarded as a random fluctuation. In the following four years (i.e., 2018-2021), the number of related papers has grown substantially, which indicates the increasing importance and popularity of RL to deal with transportation problems.

\begin{figure}[htbp]
    \centering
    \makebox[\textwidth][c]{\includegraphics[width=1.5\linewidth]{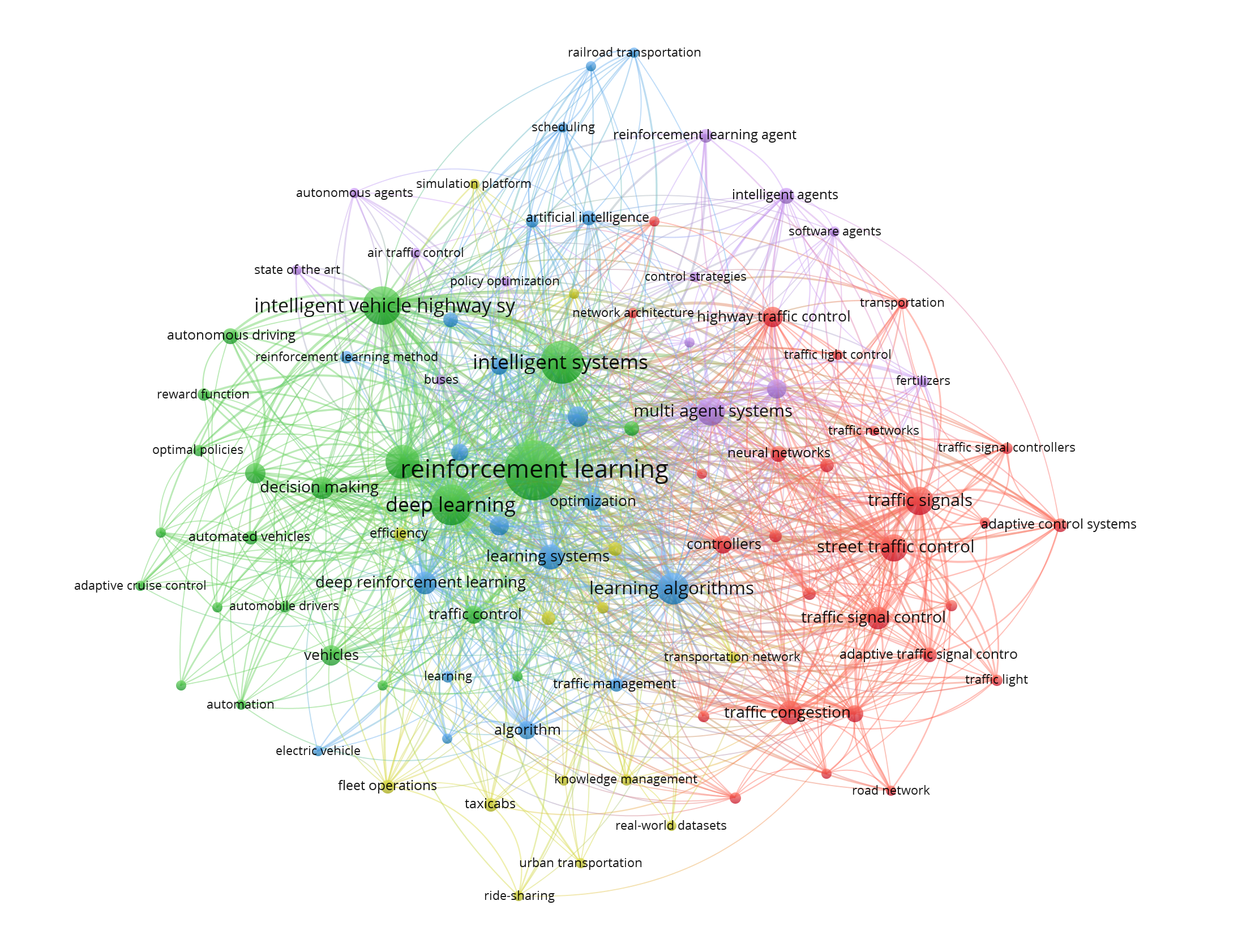}}
    \caption{Bibliographic Coupling of Keywords: the circle represents a keyword while the edge represents the co-appearance of a pair of keywords.}
    \label{fig:keyword}
\end{figure}

Furthermore, in order to identify the major transportation application areas/topics in relation to Reinforcement Learning, Fig.~\ref{fig:keyword} shows the bibliographic coupling network of keywords where the minimum number of occurrences of a keyword is five. The size of the circle represents the number of occurrences of the keyword. And the keywords represented by the same color mean the high co-appearance of these words in one paper. Excluding the words with similar meanings, the keywords with high frequency can be described as two aspects, i.e., learning algorithms and intelligent transportation applications. The learning strategies mainly cover deep learning or Neural Network and Reinforcement Learning. The major topics related to RL methods include the following nine categories: autonomous driving/vehicles, adaptive cruise control, fleet operations, ride-sharing, traffic signal control, highway/street/air traffic control, electric vehicle, taxicabs, and scheduling. Motivated by these keywords with high frequency, we identify six groups as shown in Fig.~\ref{fig:framework}, which will be reviewed in the following sections, respectively.

\section{Traffic Control: Road and Air} \label{sec:traffic_control}

Traffic control is a critical issue in traffic flow management. This section summarizes RL-based controlling strategies proposed for both roadway traffic and air traffic in order to reduce traffic congestion and delays. Due to the large number of studies for traffic signal control and to facilitate reading, we summarize studies on roadway traffic signal control (TSC) in Table~\ref{table:tsc} and summarize studies on other aspects (i.e., speed limit, price management, perimeter control, and air traffic control) in Table~\ref{table:traffic_control}.

\begin{table}[htbp]
\caption{Summary of RL Applications in Traffic Signal Control}
\begin{threeparttable}
\setlength{\tabcolsep}{.01mm}{
\begin{tabular}{c|c|c}
\hline
 &  & \textbf{Reference} \\ \hline
\multirow{8}{*}{\textbf{Framework}} & \textbf{Q-learning} & \begin{tabular}[c]{@{}c@{}}\citet{prashanth2010reinforcement}, \citet{ozan2015modified}, \\ \citet{el2013multiagent, el2014design}, \citet{mannion2015parallel}, \\ \citet{wiering2000multi}, \citet{balaji2010urban}, \citet{arel2010reinforcement} \end{tabular} \\ \cline{2-3} 
 & \textbf{DQN} & \begin{tabular}[c]{@{}c@{}}\citet{mousavi2017traffic}, \citet{wei2018intellilight}, \citet{zhang2020generalight}, \citet{xu2019targeted}, \\ \citet{van2016coordinated}, \citet{darmoul2017multi}, \citet{devailly2021igrl}, \\ \citet{wang2021adaptive}, \citet{wei2018intellilight, wei2019presslight, wei2019colight}, \citet{chen2020toward}, \\ \citet{zang2020metalight}, \citet{zhang2020using}, \citet{yu2020macar}, \citet{xu2021hierarchically} \end{tabular} \\ \cline{2-3} 
 & \textbf{A2C} & \citet{chu2019multi}, \citet{wang2021adaptive} \\ \cline{2-3} 
 & \textbf{DDPG} & \citet{li2021network}, \citet{ni2019cordon} \\ \cline{2-3} 
 & \textbf{Actor-Critic} & \citet{aslani2017adaptive} \\ \cline{2-3} 
 & \textbf{\begin{tabular}[c]{@{}c@{}}Neural fitted \\ Q-iteration\end{tabular}} & \citet{nishi2018traffic} \\ \cline{2-3} 
 & \textbf{Ape-X DQN} & \citet{zheng2019learning} \\ \hline
\multirow{2}{*}{\textbf{Agent}} & \textbf{single-agent} & \begin{tabular}[c]{@{}c@{}}\citet{prashanth2010reinforcement}, \citet{ozan2015modified}, \\ \citet{el2014design}, \citet{mousavi2017traffic}, \citet{xu2019targeted}, \\ \citet{wei2018intellilight}, \citet{zhang2020using}, \citet{ni2019cordon} \end{tabular} \\ \cline{2-3} 
 & \textbf{multi-agent} & \begin{tabular}[c]{@{}c@{}}\citet{nishi2018traffic}, \citet{wiering2000multi}, \citet{abdulhai2003reinforcement}, \\ \citet{chu2019multi}, \citet{balaji2010urban}, \citet{el2013multiagent}, \\ \citet{arel2010reinforcement}, \citet{van2016coordinated}, \citet{yu2020macar}, \\ \citet{wang2021adaptive}, \citet{zheng2019learning}, \citet{chen2020toward}, \\ \citet{xu2021hierarchically} \citet{devailly2021igrl}, \citet{mannion2015parallel}, \\ \citet{zang2020metalight}, \citet{zhang2020generalight}, \citet{wei2019presslight, wei2019colight}, \\ \citet{li2021network}, \citet{darmoul2017multi}, \citet{aslani2017adaptive} \end{tabular} \\ \hline
\multirow{2}{*}{\textbf{ \begin{tabular}[c]{@{}c@{}} Scenario/ \\ Data \end{tabular}}} & \textbf{\begin{tabular}[c]{@{}c@{}} synthetic \\ network/data \end{tabular}} & \begin{tabular}[c]{@{}c@{}}\citet{prashanth2010reinforcement}, \citet{ozan2015modified}, \\ \citet{el2014design}, \citet{mousavi2017traffic}, \citet{nishi2018traffic}, \\ \citet{wiering2000multi}, \citet{abdulhai2003reinforcement}, \citet{arel2010reinforcement}, \\ \citet{van2016coordinated}, \citet{darmoul2017multi}, \\ \citet{mannion2015parallel}, \citet{aslani2017adaptive}, \citet{ni2019cordon} \end{tabular} \\ \cline{2-3} 
 & \textbf{\begin{tabular}[c]{@{}c@{}} real-world \\ network/data \end{tabular}} & \begin{tabular}[c]{@{}c@{}} \citet{wei2018intellilight} (Jinan), \citet{zheng2019learning} (Jinan, Hangzhou), \\ \citet{zhang2020generalight} (Hangzhou, Atlanta), \\ \citet{chu2019multi} (Monaco), \citet{wang2021adaptive} (Monaco, Harbin), \\ \citet{el2013multiagent} (Toronto), \citet{li2021network} (Maryland),\\ \citet{zang2020metalight} (Jinan, Hangzhou, Atlanta, Los Angeles), \\ 
 \citet{chen2020toward, devailly2021igrl} (New York), \\ \citet{balaji2010urban} (Singapore), \citet{xu2019targeted} (Hangzhou), \\ \citet{wei2019presslight} (Jinan, New York), Zhang et al. (2020d), \\ \citet{wei2019colight, yu2020macar} (Hangzhou, Jinan, New York), \\
 \citet{xu2021hierarchically} (Hangzhou, Jinan, Shenzhen, New York) \end{tabular} \\ \hline
\multirow{8}{*}{\textbf{Simulator}} & \begin{tabular}[c]{@{}c@{}} \textbf{GLD simulator} \\ \citep{wiering2004simulation} \end{tabular}  & \citet{prashanth2010reinforcement} \\ \cline{2-3} 
 & \textbf{Paramics} & \citet{el2013multiagent, el2014design}, \citet{balaji2010urban} \\ \cline{2-3} 
 & \begin{tabular}[c]{@{}c@{}} \textbf{SUMO} \\ \citep{lopez2018microscopic}\end{tabular}
 & \begin{tabular}[c]{@{}c@{}}\citet{mousavi2017traffic}, \citet{wei2018intellilight}, \citet{nishi2018traffic}, \\ \citet{chu2019multi}, \citet{mannion2015parallel}, \citet{van2016coordinated}, \\ \citet{wang2021adaptive}, \citet{li2021network}, \citet{devailly2021igrl}, \\ \citet{zhang2020using}, \citet{yu2020macar}, \citet{xu2019targeted} \end{tabular} \\ \cline{2-3} 
 & \begin{tabular}[c]{@{}c@{}} \textbf{CityFlow} \\
 \citep{zhang2019cityflow}\end{tabular} & \begin{tabular}[c]{@{}c@{}} \citet{zhang2020generalight}, \citet{wei2019presslight, wei2019colight}, \citet{zheng2019learning}, \\ \citet{chen2020toward}, \citet{zang2020metalight}, \citet{yu2020macar}, \citet{xu2021hierarchically} \end{tabular} \\ \cline{2-3} 
 & \textbf{AIMSUN} \tnote{1} & \citet{aslani2017adaptive}, \citet{ni2019cordon} \\ \cline{2-3} 
 & \textbf{VISSIM} \tnote{2} & \citet{darmoul2017multi} \\ \cline{2-3} 
 & \textbf{\begin{tabular}[c]{@{}c@{}} personal \\ simulator \end{tabular}} & \begin{tabular}[c]{@{}c@{}} \citet{ozan2015modified}, \citet{wiering2000multi}, \\ \citet{abdulhai2003reinforcement}, \citet{arel2010reinforcement} \end{tabular} \\ \hline
\end{tabular}}
\begin{tablenotes}
\footnotesize
\item[1] http://www.AIMSUN.com
\item[2] http://vision-traffic.ptvgroup.com/en-uk/home
\end{tablenotes}
\end{threeparttable}
\label{table:tsc}
\end{table}

\begin{table}[t]
\caption{Summary of RL Applications in Speed Limit Control, Price Management, Perimeter Control, and Air Traffic Control}
\begin{threeparttable}
\setlength{\tabcolsep}{.2mm}{
\begin{tabular}{c|c|c|c|c|c}
\hline
\textbf{Reference} & \textbf{Application} & \textbf{Framework} & \textbf{Agent} & \textbf{Scenario/Data} & \textbf{Simulator} \\ \hline
\begin{tabular}[c]{@{}c@{}} \citet{zhu2014accounting} \end{tabular} & \begin{tabular}[c]{@{}c@{}} speed limit \\ control \end{tabular} & TD-based RL & \begin{tabular}[c]{@{}c@{}}single-agent, \\ the controller\end{tabular} & \begin{tabular}[c]{@{}c@{}} Sioux Falls \\ network \end{tabular} & \begin{tabular}[c]{@{}c@{}}personal \\ simulator\end{tabular} \\ \hline
\citet{li2017reinforcement} & \begin{tabular}[c]{@{}c@{}} speed limit \\ control \end{tabular} & Q-learning & \begin{tabular}[c]{@{}c@{}}single-agent, \\ the controller\end{tabular} & \begin{tabular}[c]{@{}c@{}}Interstate \\ freeway \\ in Oakland \end{tabular} & \begin{tabular}[c]{@{}c@{}}personal \\ simulator\end{tabular} \\ \hline
\citet{wu2020differential} & \begin{tabular}[c]{@{}c@{}} speed limit \\ control \end{tabular} & DDPG & \begin{tabular}[c]{@{}c@{}}single-agent, \\ the controller\end{tabular} & \begin{tabular}[c]{@{}c@{}}northbound \\ freeway of I405 \\ in California \end{tabular} & SUMO \\ \hline 
\citet{pandey2018multiagent} & \begin{tabular}[c]{@{}c@{}} price \\ management \end{tabular} & \begin{tabular}[c]{@{}c@{}}Sparse \\ Cooperative \\ Q-learning\end{tabular} & \begin{tabular}[c]{@{}c@{}}multi-agent, \\ a toll\end{tabular} & \begin{tabular}[c]{@{}c@{}} synthetic \\ network \end{tabular} & \begin{tabular}[c]{@{}c@{}}personal \\ simulator\end{tabular} \\ \hline
\citet{pandey2020deep} & \begin{tabular}[c]{@{}c@{}} price \\ management \end{tabular} & A2C, PPO & \begin{tabular}[c]{@{}c@{}}multi-agent, \\ a toll\end{tabular} & \begin{tabular}[c]{@{}c@{}}express lanes \\ in Dallas \\ and Austin\end{tabular} & \begin{tabular}[c]{@{}c@{}}personal \\ simulator\end{tabular} \\ \hline
\citet{zhou2021model} & \begin{tabular}[c]{@{}c@{}} perimeter \\ control \end{tabular} & DQN, DDPG & \begin{tabular}[c]{@{}c@{}}single-agent, \\ the controller\end{tabular} & \begin{tabular}[c]{@{}c@{}} synthetic \\ network \end{tabular} & \begin{tabular}[c]{@{}c@{}}personal \\ simulator\end{tabular} \\ \hline
\citet{chen2022data} & \begin{tabular}[c]{@{}c@{}} perimeter \\ control \end{tabular} & \begin{tabular}[c]{@{}c@{}} Policy \\ iteration \end{tabular} & \begin{tabular}[c]{@{}c@{}}single-agent, \\ the controller\end{tabular} & \begin{tabular}[c]{@{}c@{}} synthetic \\ network \end{tabular} & SUMO \\ \hline
\citet{yang2017multi} & \begin{tabular}[c]{@{}c@{}} perimeter \\ control \end{tabular} & DQN & \begin{tabular}[c]{@{}c@{}}single-agent, \\ the controller\end{tabular} & \begin{tabular}[c]{@{}c@{}} synthetic \\ network \end{tabular} & \begin{tabular}[c]{@{}c@{}}personal \\ simulator\end{tabular} \\ \hline
\citet{rezaee2012application} & \begin{tabular}[c]{@{}c@{}} ramp \\ metering \end{tabular} & Q-learning & \begin{tabular}[c]{@{}c@{}}single-agent, \\ the controller\end{tabular} & \begin{tabular}[c]{@{}c@{}} in the City \\ of Toronto \end{tabular} & Paramics \\ \hline
\citet{fares2014freeway} & \begin{tabular}[c]{@{}c@{}} ramp \\ metering \end{tabular}  & Q-learning & \begin{tabular}[c]{@{}c@{}}single-agent, \\ the controller\end{tabular} & \begin{tabular}[c]{@{}c@{}} synthetic \\ network \end{tabular} & \begin{tabular}[c]{@{}c@{}}personal \\ simulator\end{tabular} \\ \hline
\citet{belletti2017expert} & \begin{tabular}[c]{@{}c@{}} ramp \\ metering \end{tabular}  & DDPG & \begin{tabular}[c]{@{}c@{}}multi-agent, \\ the controller \\ for a region\end{tabular} & \begin{tabular}[c]{@{}c@{}}San Francisco \\ Bay Bridge\end{tabular} & BeATs \tnote{1} \\ \hline
\citet{tumer2007distributed} & \begin{tabular}[c]{@{}c@{}}air traffic \\ management\end{tabular} & Q-learning & \begin{tabular}[c]{@{}c@{}}multi-agent, \\ a location\end{tabular} &  \begin{tabular}[c]{@{}c@{}} synthetic \\ network \end{tabular} & FACET \tnote{2} \\ \hline
\citet{balakrishna2010accuracy} & flight delay & Q-learning & \begin{tabular}[c]{@{}c@{}}single-agent, \\ the controller\end{tabular} & \begin{tabular}[c]{@{}c@{}}Tampa \\ International \\ Airport\end{tabular} & \begin{tabular}[c]{@{}c@{}}personal \\ simulator\end{tabular} \\ \hline
\end{tabular}}
\begin{tablenotes}
\footnotesize
\item[1] https://connected-corridors.berkeley.edu/berkeley-advanced-traffic-simulator
\item[2] https://www.nasa.gov/centers/ames/research/lifeonearth/lifeonearth-facet.html
\end{tablenotes}
\end{threeparttable}
\label{table:traffic_control}
\end{table}

\subsection{Roadway Traffic Control}

On roadway traffic control, we review the following four major issues: traffic signal control; speed limit control; pricing management; perimeter control; and ramp metering.

\subsubsection{Traffic Signal Control}

\begin{figure}[htb]
\centering
\begin{subfigure}{.4\textwidth}
\includegraphics[width=\linewidth]{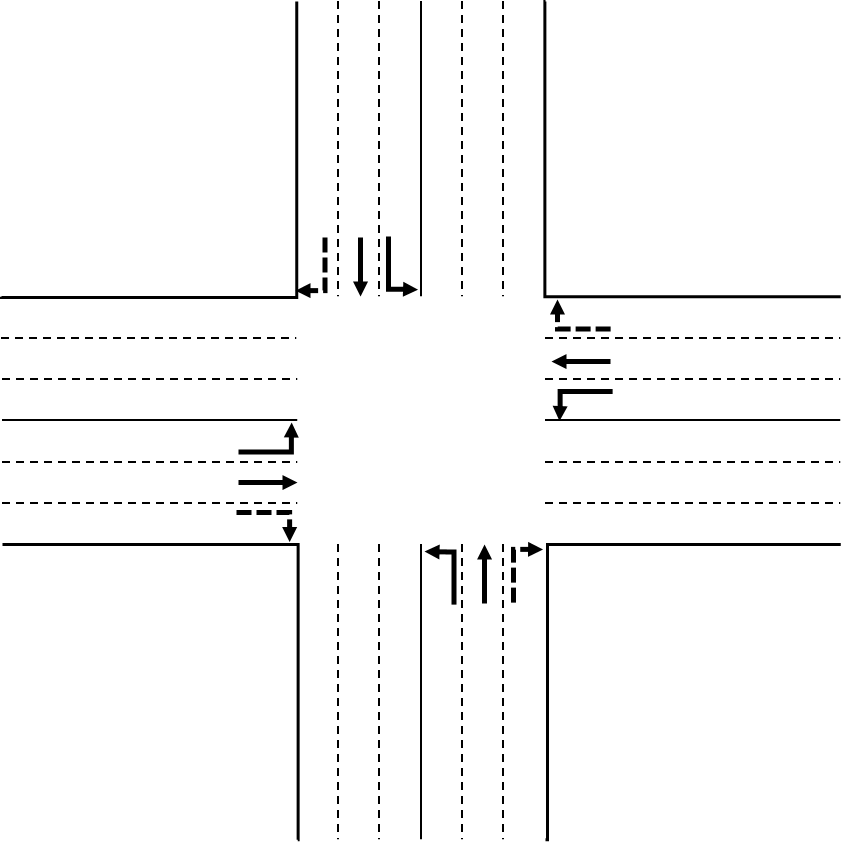}
\caption{A Typical Four-Approach Intersection}
\label{fig:intersection}
\end{subfigure}
\hspace{1cm}
\begin{subfigure}{.4\textwidth}
\includegraphics[width=\linewidth]{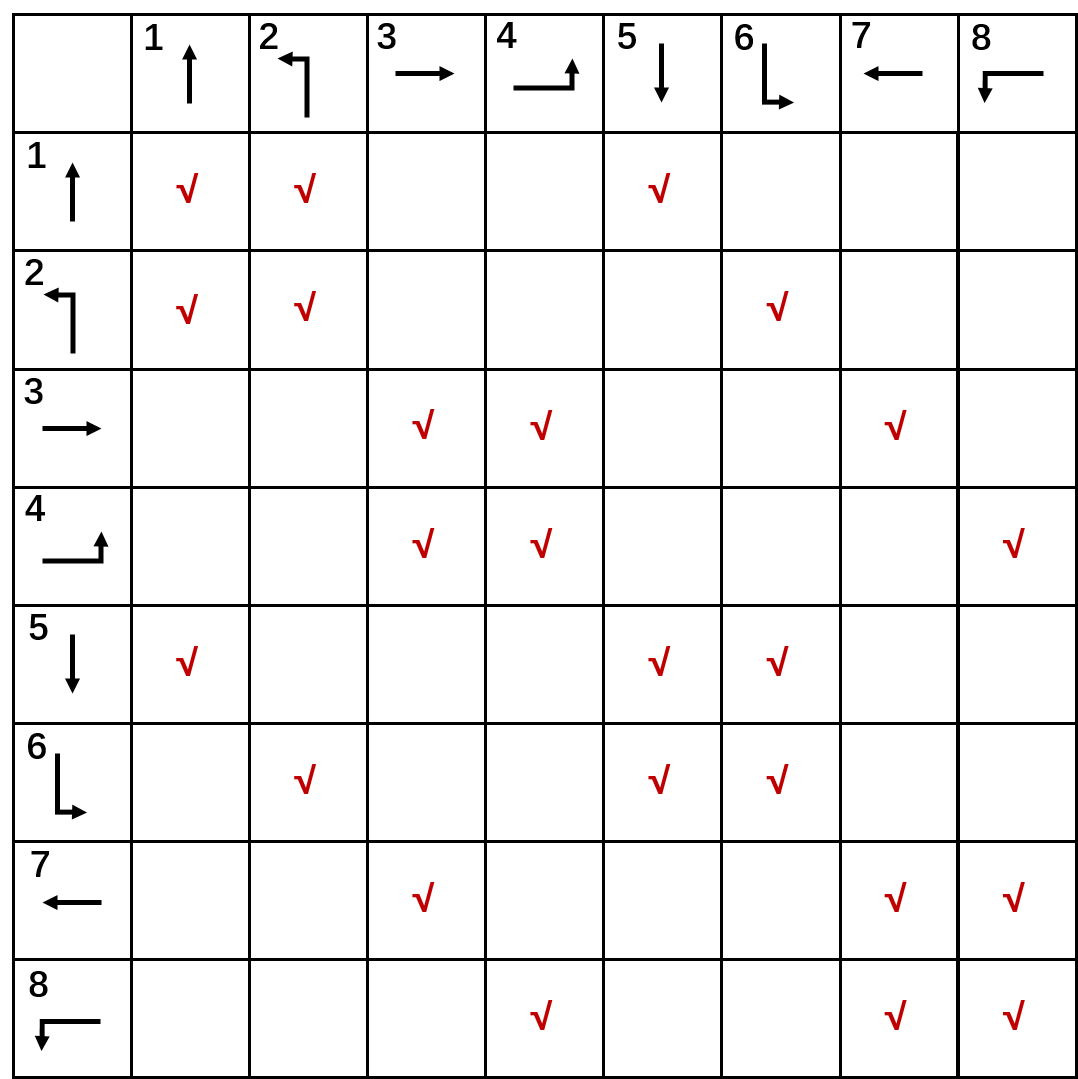}
\caption{Phase Competition Matrix}
\label{fig:matrix}
\end{subfigure}
\caption{Traffic Signal Related Schematic Diagrams}
\label{fig:singal}
\end{figure}

The congestion and delays caused by traffic bottlenecks motivate the development of methods for traffic signal control (TSC) \citep{yau2017survey}. Conventional pre-timed control systems set constant time signals, while RL-based approaches have been used to dynamically and adaptively optimize traffic signal timing. An illustration of a four-approach intersection is provided in Fig.~\ref{fig:intersection} (left-hand driving is assumed), and a typical signal plan with eight phases is shown in Fig.~\ref{fig:singal}. Many studies are formulated based on the four-approach intersections with eight phases \citep{arel2010reinforcement}.

Earlier studies only address the traffic signal control for one intersection with single-agent RL methods, which are hard to be adaptable in environments with multiple intersections. Specifically, for one intersection, an intuitive designing scheme is to regard the intersection as an agent for signal control policy optimization. In RL methods, the agent's decision is subject to the setting of phases. To deal with the single-agent (one intersection) scenario, Q-learning has been directly adopted as the framework to learn the action-value \citep{prashanth2010reinforcement, ozan2015modified, el2014design} in order to reduce the total/average delay of vehicles. To further reduce the dimensionality of the state, \citet{prashanth2010reinforcement} takes the queue lengths and elapsed times of each signaled lane as indicators for the level of congestion (i.e., low, medium, or high), and the congestion level is used as the state. \citet{ozan2015modified} and \citet{el2014design} utilize exact values regarding traffic conditions (e.g., link flows and the free-flow travel time) as the state when optimizing the cycle time, which can be more practical than level settings in previous studies. Similarly, \citet{mousavi2017traffic} views the intersection as a vector of row pixel values, and \citet{wei2018intellilight} uses the image representation of vehicles’ positions as the state. And these studies \citep{mousavi2017traffic, wei2018intellilight, zhang2020using} then adopt the deep model, DQN, for traffic light optimization that accommodates complex and non-linear environmental information of an intersection.

The control strategies for one intersection can hardly relieve the traffic congestion in large metropolis with complex and dense networks, which motivates the traffic control studies to simultaneously consider multiple intersections. As multiple intersections (especially neighboring intersections) may interact with each other, the optimal policy strategies should be considered at the target-area level to further improve traffic efficiency. Many studies have set the reward function as the overall waiting time \citep{wiering2000multi, nishi2018traffic}, overall delay \citep{abdulhai2003reinforcement, balaji2010urban} of all vehicles in multiple intersections, or the pressure \citep{varaiya2013max} of all intersections \citep{wei2019presslight}. Though these studies achieve satisfactory performance, the relations or impacts among various intersections have not been explored explicitly.

Followed by the multi-intersection research, a series of studies focus on the coordination or competition among multiple agents/intersections to find area-wide or system-wide TSC strategies. \citet{el2013multiagent} adopts the principle of Multi-agent Modular Q-learning \citep{ono1996multi} to explicitly analyze the correlations of the target agent and one of its neighbor intersections to learn the joint policy. \citet{arel2010reinforcement} designs two types of agents for collaboration, a central agent extracting the information from itself and neighboring intersections to further learn a value function and an outbound agent to schedule its own signals. And Q-learning is used as the optimizing strategy. Furthermore, based on Advantage Actor-Critic (A2C), \citet{chu2019multi} constructs the state of the agent as the composition of its observation and neighbor policies to achieve agents' coordination. The performance of the discussed coordination-based methods is superior to the isolated intersection models in terms of average intersection delay, queue length, link stop time, and link travel time.

In the aforementioned approaches, the agent of an intersection communicates with its adjacent locations but does not coordinate with further away intersections. To address more general system-wide or area-wide issues, adopting DQN as the basic framework, \citet{van2016coordinated} combines multiple local Q-functions linearly as a global Q-function and utilizes the max-plus coordination algorithm \citep{kok2005using} to optimize the joint action for multiple intersections in an area. Similarly, \citet{mannion2015parallel} defines Master and Slave agents where the Master agent uses a shared experience pool to deal with experiences from Master Agents for coordination. \citet{yu2020macar} designs an active cross-agent communication mechanism to generate coordinated actions and uses the predicted traffic of the whole road network to mitigate the unnecessary impact of other agents’ actions. Moreover, in \citet{wang2021adaptive}, the Mobile Edge Computing (MEC) server with a fixed number of Road Side Units (RSUs) collects and deals with the local states from target intersections. The processed information is sent back to each individual agent to decide the phase of the traffic light. \citet{li2021network} proposes a shared knowledge container to store the information obtained from the whole environment by embedding the observation vectors through Gated Recurrent Unit (GRU). Each agent then chooses relevant features from the container to make its own decision based on the Deep Deterministic Policy Gradient (DDPG) algorithm.

The aforementioned studies test their approaches on small-scale environments for illustration (e.g., one intersection or dozens of intersections) while leaving scalability issues and large-scale applications for further research. In practice, megalopolis are usually installed with thousands of traffic light intersections, which need to be controlled simultaneously. Thus, to handle large-scale problems in recent years, \citet{wei2019colight} designs a graph attentional network for agents' coordination by calculating and normalizing the importance score (i.e., the value to evaluate the importance of the information from the source intersection when determining the policy for the target intersection) for all intersections in pairs. The influence affected by relevant intersections is modeled by the combination of the representation of the target agent and its corresponding importance score. However, determining the importance score in pair still occupies a large number of computation resources. To reduce the exploration space, \citet{zheng2019learning} proposes the FRAP (i.e., Flipping and Rotation and considers All Phase configurations) model to calculate the phase score in the control problem. The in-variance to symmetries (e.g., flipping and rotation) in traffic signal control is achieved by pair-wise phase completion modeling to reduce the exploration space under complex scenarios. The score of the target phase is obtained by the element-wise multiplication of the phase pair demand representation and the phase competition mask. The representation is obtained by the number of vehicles and the current signal phase, and the mask is derived from the phase competition matrix shown in Fig.~\ref{fig:matrix}. The phase with the highest score is chosen to be the action. The method is combined with both value-based and policy-based RL algorithms for optimization. Furthermore, \citet{chen2020toward} combines PressLight \citep{wei2019presslight} for reward function designing and FRAP \citep{zheng2019learning} for faster training process with parameter sharing among the agents. The model is evaluated on a simulated environment with thousands of intersections to show its effectiveness. More recently, \citet{xu2021hierarchically} illustrates that minimizing the queue length, waiting time, or delay is not equivalent to minimizing average travel time, which motivates the design of different agents with different optimizing sub-targets (e.g., queue length). A high-level policy is then proposed to align all sub-policies and avoid directly minimizing average travel time.

The optimization for large-scale environments needs numerous computational resources and time, which limits such strategies to be put into practice. Therefore, given that insufficient relevant data or computing resources in the target area, \citet{xu2019targeted, zang2020metalight, zhang2020generalight, devailly2021igrl} propose to transfer and adapt experiences learned from existing scenarios to new scenarios, which can reduce the reliance on sufficient data and decrease training consumption. As for the transfer strategies, \citet{xu2019targeted} selects similar source and target intersections by calculating similarity values, \citet{zang2020metalight, zhang2020generalight} adopt Meta-Reinforcement Learning \citep{finn2018meta}, while \citet{devailly2021igrl} applies zero-short transfer learning \citep{higgins2017darla} into the TSC framework. As for the framework of Reinforcement Learning, \citet{zang2020metalight} develops a model based on FRAP \citep{zheng2019learning} and \citet{xu2019targeted, zhang2020generalight, devailly2021igrl} utilize DQN directly.

The aforementioned studies focus on regular traffic situations while \citet{darmoul2017multi, aslani2017adaptive} focus on finding optimal solutions for traffic disruptions that are also practical and useful. In detail, \citet{darmoul2017multi} investigates the impact of accidents on traffic light control by mitigating the concepts of primary and secondary immune responses (i.e., the disturbance on the road is regarded as an antigen and the associated control decision is denoted as an antibody). The multi-agent DQN method has been used for policy optimization. More specifically, the studied traffic network in \citet{aslani2017adaptive} considers impatient pedestrians with illegal crossing behavior, vehicles parking beside the streets, and incidents (e.g., vehicle breakdown). The Actor-Critic framework is adopted to determine the duration of each phase (red/green light), which shows the capability of reducing average travel time when traffic disruptions have occurred. Furthermore, cordon control to determine the traffic signal metering rates is also an efficient way for vehicle inflows restriction. To find the optimal distribution for the metered vertices of roads, \citet{ni2019cordon} adopts the Graph Convolution Network (GCN) to formulate the directed graph representation of the environment (i.e., the street network’s geometry) and traffic (i.e., traffic conditions and directions of movements) of an intersection. The optimal actions are obtained via the DDPG method to maximize the metered flow passing through the cordon.

The satisfactory performance on traffic signal control with Reinforcement Learning provides a novel and useful tool for other transportation control issues, which motivates the usage of RL in transportation applications in recent ten years.

\subsubsection{Speed Limit Control}

For flow maximization, speed limit control (adjusting the speed limit) is often used to drive the freeway recurrent traffic bottleneck density to be close to the desired density and thus avoid capacity drops \citep{liu2015effectiveness}. The mechanism of conventional feedback-based strategies requires significant time \citep{li2017reinforcement}, which stimulates adopting RL-based methods to deal with highly dynamic traffic situations in a timely manner.

Research for speed limit control has evolved from discrete formulations to continuous formulations in order to meet complex and varying environments. Specifically, a speed limit controller is designed in \citet{zhu2014accounting} to work as the agent. Four congestion levels (i.e., free flow state, slight congestion state, moderate congestion state, and heavy congestion state) are defined as the input state of the model based on the flow density. The policy is optimized by the temporal difference (TD) algorithm. However, four discrete congestion levels might not be sufficient to fully depict the complicated and varying environment that would affect decision-making. Thus, \citet{li2017reinforcement} uses the density at the downstream of the merge area, the density at the upstream mainline section, and the density on the ramp by specific variables instead of congestion levels to minimize the travel time. The posted speed limits set as integer multiples of five mph for freeway bottlenecks are determined by the Q-learning strategy. Similar state representations are utilized in \citet{wu2020differential} for variable speed limits control based on the optimization by the DDPG algorithm with single-agent. The proposed method is able to reduce congestion, accidents, and emissions by defining the reward function as the combination of total travel time, average velocity reported by detectors, the number of emergency braking vehicles, and related gas emissions. Though the research for speed limit control with RL does not receive much attention, the success of existing studies provides a solid foundation for future optimization.

\subsubsection{Pricing}

Dynamic pricing for managed lanes can be used to offer a premium service and alleviate congestion \citep{devarasetty2014can}. \citet{pandey2018multiagent} and \citet{pandey2020deep} examine pricing management via Reinforcement Learning to find optimal policies that maximize the revenue of the managed lanes. In these strategies, the vector containing the number of vehicles detected by the loop detectors is used as the state while the toll is set as the agent at the entrance of each managed line to decide the real-time price. A sparse cooperative Q-learning algorithm \citep{kok2006collaborative} is adopted in \citet{pandey2018multiagent} while A2C and PPO are used in \citet{pandey2020deep} to optimize the pricing policy.

\subsubsection{Perimeter Control}

Perimeter control is regarded as an efficient way for regional traffic control that can benefit all users \citep{yang2017multi}. The appealing performance obtained by RL-based optimizing strategies for traffic signal control illustrates their ability to handle complex and varying road environments. Similar environments analyzing in perimeter control and traffic signal control provide a novel direction for perimeter control, i.e., RL-based methods. Specifically, in \citet{yoon2020design}, the agent determines green time ratios as discrete values with the optimization by DQN. However, this method is only able to handle discrete actions, which is less practical. To avoid relying on the full knowledge of the road network and design continuous action, \citet{zhou2021model-conference, zhou2021model} proposes an RL-based scheme for an urban network composed of two homogeneous sub-regions to improve the network throughput (i.e., the number of trips completed). Discrete-RL (D-RL) model optimized by DQN and Continuous-RL (C-RL) model optimized by DDPG are designed for discrete actions and continuous actions, respectively. Acknowledging the information of accumulations and estimated traffic demands as the state, the agent of D-RL decides the range while the agent of C-RL controls the allowable decrease/increase value of perimeter controllers (i.e., the parameter defined by the allowable portions of transfer flows) by maximizing actual portions of transfer flows. In addition, \citet{chen2022data} proposes a deep-based integral policy iteration approach to minimize the total time spent for multi-region perimeter control in the continuous manner.

\subsubsection{Ramp Metering}

Ramp metering takes advantage of traffic signals at freeway on-ramps to control the rate of vehicles entering the freeway. To decide passing and prohibiting phases on the freeway, the information of the numbers of vehicles in the mainstream and entering the freeway and the status of the ramp traffic signal are denoted as the state. \citet{rezaee2012application} and \citet{fares2014freeway} utilize Q-learning-based methods to minimize the total travel time of the whole network and the freeway density, respectively. The proposed models have been tested on a case study (e.g., the City of Toronto) and a synthetic network, which illustrates the effectiveness of RL-based methods in dealing with the ramp metering problem. However, the aforementioned two single-agent-based methods have limited scalability for controlling numerous intersections simultaneously. This motivates \citet{belletti2017expert} to design a multi-agent DDPG framework for ramp metering. The highway vehicle density is modeled by the Partial Differential Equation to decide the incoming flow by maximizing the total observed outflow with the policy gradient algorithm. The interaction among agents is achieved by the introduction of Mutual Weight Regularization \citep{caruana1997multitask}.

\subsection{Air Traffic Control}

Congestion in air traffic creates substantial flight delays and limits efficiency and productivity. As reported in \citet{balakrishna2010accuracy}, one of the major factors leading to flight delays is the taxi-out delay (i.e., the time between gate push back and time of takeoff). In order to mitigate congestion in the airport, a novel way to predict the delay based on RL is proposed, which has a relatively low demand on training data for optimization when compared to classical supervised learning strategies. The agent learns the information from the environment of the aircraft and airport (e.g., the number of aircraft in the queue at the runway and the number of departure aircraft co-taxiing) to estimate the taxi-out time by minimizing the absolute value of the error between the actual taxi-out time and predicted taxi-out time. In addition, \citet{tumer2007distributed} applies multi-agent Reinforcement Learning in air traffic flow management to minimize the sum of total delay penalty and total congestion penalty for all aircraft in the system. The ground locations throughout the airspace are split into multiple individual `fixes' (i.e., individual locations) where each `fix' is regarded as an agent. The task of the agent is to decide the distance between the approaching aircraft and itself, which can control the rate of aircraft going through a `fix'. The proposed method is tested on a simulation tool, FACET, developed by NASA to show its ability for congestion reduction. The effectiveness of numerous RL strategies for air traffic control still needs to be conducted and evaluated in future research with more complicated and practical scenarios.

\section{Taxi and Ride-sourcing/sharing} \label{sec:taxi}

Cooperative mobility-on-demand (MOD) systems (e.g., Uber, Lyft, and Didi Chuxing) have been spreading widely \citep{he2019spatio} and provide multiple online taxi services such as express car, ride-sharing, ride-sourcing, and traditional taxi. The real-time large-scale order information provides the opportunity to analyze demand patterns for further forecasting and management. To reduce resource utilization, decrease the waiting time, and increase profit, Reinforcement Learning has been investigated for vehicle re-positioning, order dispatching, and vehicle routing in the taxi and ride-sourcing/sharing service systems, where a summary of related papers is provided in Table~\ref{table:taxi}.

\begin{table}[]
\caption{Summary of RL Applications in Taxi and Ride-Sourcing/Sharing Service Systems}
\begin{threeparttable}
\setlength{\tabcolsep}{.01mm}{
\begin{tabular}{c|c|c|c|c|c}
\hline
\textbf{Reference} & \textbf{Application} & \textbf{Framework} & \textbf{Agent} & \textbf{Data} & \textbf{Simulator} \\ \hline
\citet{lin2018efficient} & \begin{tabular}[c]{@{}c@{}}vehicle re-positioning \\ for ride-hailing systems\end{tabular} & \begin{tabular}[c]{@{}c@{}}Contextual \\ DQN and \\ Actor-Critic\end{tabular} & \begin{tabular}[c]{@{}c@{}}multi-agent, \\ an available \\ vehicle\end{tabular} & \begin{tabular}[c]{@{}c@{}}real data from \\ Didi Chuxing \\ in Chengdu\end{tabular} & \begin{tabular}[c]{@{}c@{}} contextual \\ simulator \\ \citep{lin2018efficient} \end{tabular} \\ \hline
\citet{shou2020reward} & \begin{tabular}[c]{@{}c@{}}vehicle re-positioning \\ for traditional \\ taxi systems\end{tabular} & \begin{tabular}[c]{@{}c@{}}Mean Field \\ Actor-Critic \\ algorithm\end{tabular} & \begin{tabular}[c]{@{}c@{}}multi-agent, \\ an available \\ vehicle\end{tabular} & \begin{tabular}[c]{@{}c@{}}synthetic data, \\ real data \\ from NYC TLC \tnote{1} \end{tabular} & \begin{tabular}[c]{@{}c@{}} personal \\ simulator \end{tabular} \\ \hline
\citet{nguyen2017policy}& \begin{tabular}[c]{@{}c@{}}vehicle re-positioning \\ for traditional \\ taxi systems\end{tabular} & \begin{tabular}[c]{@{}c@{}} Actor-Critic \\ algorithm \end{tabular} & \begin{tabular}[c]{@{}c@{}}multi-agent, \\ an available \\ vehicle\end{tabular} & \begin{tabular}[c]{@{}c@{}}synthetic data, \\ real taxi data \\ from Singapore\end{tabular} & \begin{tabular}[c]{@{}c@{}} personal \\ simulator \end{tabular} \\ \hline
\citet{mao2020dispatch} & \begin{tabular}[c]{@{}c@{}}vehicle re-positioning \\ for traditional \\ taxi systems \end{tabular} & \begin{tabular}[c]{@{}c@{}} Deep \\ Actor-Critic \\ algorithm \end{tabular} & \begin{tabular}[c]{@{}c@{}}multi-agent, \\ an available \\ vehicle\end{tabular} & \begin{tabular}[c]{@{}c@{}} real data \\ from NYC TLC \tnote{1}\end{tabular} & \begin{tabular}[c]{@{}c@{}} personal \\ simulator \end{tabular} \\ \hline
\citet{oda2018movi} & \begin{tabular}[c]{@{}c@{}}order dispatching\\ for traditional \\ taxi systems\end{tabular} & Double-DQN & \begin{tabular}[c]{@{}c@{}}single-agent, \\ dispatch \\ center\end{tabular} & \begin{tabular}[c]{@{}c@{}} real data \\ from NYC TLC \tnote{1}\end{tabular} & \begin{tabular}[c]{@{}c@{}} personal \\ simulator \end{tabular} \\ \hline
\citet{zhou2019multi} & \begin{tabular}[c]{@{}c@{}}order dispatching \\ for ride-hailing \\ systems\end{tabular} & DQN & \begin{tabular}[c]{@{}c@{}}multi-agent,\\ a driver\end{tabular} & \begin{tabular}[c]{@{}c@{}}real data from \\ Didi Chuxing of \\ three cities\end{tabular} & \begin{tabular}[c]{@{}c@{}} simulator \\ provided by \\ Didi Chuxing \end{tabular} \\ \hline
\citet{xu2018large} & \begin{tabular}[c]{@{}c@{}}order dispatching \\ for ride-hailing systems\end{tabular} & TD-based RL & \begin{tabular}[c]{@{}c@{}}multi-agent,\\ a driver\end{tabular} & \begin{tabular}[c]{@{}c@{}}synthetic data, \\ real data from \\ Didi Chuxing \end{tabular} & \begin{tabular}[c]{@{}c@{}} personal \\ simulator \end{tabular} \\ \hline
\citet{li2019efficient} & \begin{tabular}[c]{@{}c@{}}order dispatching \\ for ride-hailing systems\end{tabular} & \begin{tabular}[c]{@{}c@{}}Actor-Critic,\\ Mean Field RL\end{tabular} & \begin{tabular}[c]{@{}c@{}}multi-agent,\\ a driver\end{tabular} & \begin{tabular}[c]{@{}c@{}}real data from\\ Didi Chuxing\end{tabular} & \begin{tabular}[c]{@{}c@{}} contextual \\ simulator \\ \citep{lin2018efficient} \end{tabular} \\ \hline
\citet{he2019spatio} & order dispatching & Double-DQN & \begin{tabular}[c]{@{}c@{}}single-agent, \\ coordination \\ center\end{tabular} & \begin{tabular}[c]{@{}c@{}}real data from \\ Uber, Yellow Taxi \\ and Didi Chuxing \end{tabular} & \begin{tabular}[c]{@{}c@{}} personal \\ simulator \end{tabular} \\ \hline
\citet{wang2018deep} & \begin{tabular}[c]{@{}c@{}}order dispatching \\ for ride-hailing systems\end{tabular} & Double-DQN & \begin{tabular}[c]{@{}c@{}}multi-agent,\\ a driver\end{tabular} & \begin{tabular}[c]{@{}c@{}}ExpressCar data \\ from Didi Chuxing\end{tabular} & \begin{tabular}[c]{@{}c@{}} personal \\ simulator \end{tabular} \\ \hline
\citet{tang2019deep} & \begin{tabular}[c]{@{}c@{}}order dispatching \\ for ride-hailing systems\end{tabular} & TD-based RL & \begin{tabular}[c]{@{}c@{}}multi-agent,\\ a driver\end{tabular} & \begin{tabular}[c]{@{}c@{}}real data from\\ Didi Chuxing\end{tabular} & \begin{tabular}[c]{@{}c@{}} personal \\ simulator \end{tabular} \\ \hline
\citet{jin2019coride} & \begin{tabular}[c]{@{}c@{}}order dispatching and \\ vehicle re-position\\ for ride-hailing systems\end{tabular} & \begin{tabular}[c]{@{}c@{}}Hierarchical \\ RL, DDPG\end{tabular} & \begin{tabular}[c]{@{}c@{}}multi-agent,\\ a region cell\end{tabular} & \begin{tabular}[c]{@{}c@{}}real data from \\ Didi Chuxing \end{tabular} & \begin{tabular}[c]{@{}c@{}} contextual \\ simulator \\ \citep{lin2018efficient} \end{tabular} \\ \hline
\citet{holler2019deep} & \begin{tabular}[c]{@{}c@{}}order dispatching and \\ vehicle re-position\\ for ride-hailing systems\end{tabular} & DQN, PPO & \begin{tabular}[c]{@{}c@{}}multi-agent,\\ a driver\end{tabular} & \begin{tabular}[c]{@{}c@{}}synthetic data, \\ real GAIA dataset\\ from Didi Chuxing\end{tabular} & \begin{tabular}[c]{@{}c@{}} personal \\ simulator \end{tabular} \\ \hline
\citet{chen2019inbede} & \begin{tabular}[c]{@{}c@{}}order dispatching and \\ pricing for \\ ride-hailing systems\end{tabular} & TD-based RL & \begin{tabular}[c]{@{}c@{}}single-agent, \\ coordination \\ center\end{tabular} & \begin{tabular}[c]{@{}c@{}}real data from \\ Didi Chuxing \end{tabular} & \begin{tabular}[c]{@{}c@{}} simulator \\ provided by \\ Didi Chuxing \end{tabular} \\ \hline
\citet{manchella2021flexpool} & \begin{tabular}[c]{@{}c@{}}order dispatching and \\ goods delivery for \\ ride-hailing systems \end{tabular} & Double-DQN & \begin{tabular}[c]{@{}c@{}}multi-agent, \\ a vehicle \end{tabular} & \begin{tabular}[c]{@{}c@{}}real data from \\ New York City \\ Taxicab\end{tabular} & \begin{tabular}[c]{@{}c@{}} personal \\ simulator \end{tabular} \\ \hline
\citet{james2019online} & \begin{tabular}[c]{@{}c@{}}vehicle routing for \\ ride-hailing systems\end{tabular} & \begin{tabular}[c]{@{}c@{}}Deep Policy \\ Gradient \\ algorithm \end{tabular} & \begin{tabular}[c]{@{}c@{}}single-agent, \\ dispatch \\  center\end{tabular} & \begin{tabular}[c]{@{}c@{}}real data from \\ Cologne \end{tabular} & \begin{tabular}[c]{@{}c@{}} personal \\ simulator \end{tabular} \\ \hline
\citet{zhang2020multi} & \begin{tabular}[c]{@{}c@{}}vehicle routing for \\ ride-hailing systems\end{tabular} & \begin{tabular}[c]{@{}c@{}}Deep Policy \\ Gradient \\ algorithm\end{tabular} & \begin{tabular}[c]{@{}c@{}}multi-agent,\\ a vehicle\end{tabular} & synthetic data & \begin{tabular}[c]{@{}c@{}} personal \\ simulator \end{tabular} \\ \hline
\citet{al2019deeppool} & \begin{tabular}[c]{@{}c@{}}order dispatching and \\ vehicle routing for \\ ride-hailing systems\end{tabular} & Double-DQN & \begin{tabular}[c]{@{}c@{}}single-agent, \\ dispatch \\ center\end{tabular} & \begin{tabular}[c]{@{}c@{}} real data of taxi \\ from NYC TLC \tnote{1} \end{tabular} & \begin{tabular}[c]{@{}c@{}} personal \\ simulator \end{tabular} \\ \hline
\end{tabular}}
\begin{tablenotes}
\footnotesize
\item[1] https://www1.nyc.gov/site/tlc/about/tlc-trip-record-data.page
\end{tablenotes}
\end{threeparttable}
\label{table:taxi}
\end{table}

\subsection{Vehicle Re-positioning}

The imbalance between supply and demand leads to long waiting times for passengers and time/energy loss for drivers. Re-positioning available vehicles/drivers to potential locations (e.g., locations with massive demand) is necessary to improve system efficiency and better match supply and demand. Methods requiring accurate information on a wide range of parameters or variables (e.g., customer demand and travel time) are often time-consuming \citep{mao2020dispatch}. Therefore, RL-based methods without the need for prior knowledge are broadly utilized for vehicle re-positioning in taxi and ride-sourcing/sharing systems.

In the ride-hailing system, considering the influence among all vehicles and customers, existing studies \citep{nguyen2017policy, lin2018efficient, shou2020reward, mao2020dispatch} take each available vehicle (or driver) as an agent for vehicle re-position with multi-agent RL models. In detail, \citet{lin2018efficient} applies a multi-agent RL-based strategy to maximize gross merchandise volume (GMV, i.e., the number of all orders served) and improve order response rate. A contextual DQN model is designed for the allocation instructing, including the geographic context to filter out invalid directions for agents and the collaborative context to avoid agents moving in conflicting directions. And a contextual Actor-Critic framework is designed for explicit coordination among agents to enhance policy-making by acknowledging spatial distributions of available vehicles and orders. As for the traditional taxi system, global information such as the distribution of all taxis is hard to be obtained in a short time for optimization. Thus, \citet{shou2020reward} develops a taxi re-positioning method that only uses local observations from each driver/vehicle through multi-agent Mean Field Actor-Critic algorithm \citep{yang2018mean}. The aim of each agent (i.e., an available vehicle/driver) is to maximize their own monetary return. To accommodate the selfishness of each agent, Bayesian optimization is adopted to design the reward function, which helps achieve a better equilibrium for the overall system. 

The computational complexity of the vanilla Actor-Critic-based method is relatively high for large-scale multi-agent vehicle re-positioning, which can take a very long time for convergence and is neglected in \citet{lin2018efficient, shou2020reward}. Thus, in favor of reducing the computational complexity and speeding up the optimization process, \citet{nguyen2017policy} decomposes the approximation of the action-value function over agents and derives a modified loss function to train the critic for each agent based on its own reward. The proposed strategy is tested on datasets with a large agent population size to decide whether drivers should stay in the current zone or move to another zone to look for passengers for total profit maximization.

The influence of waiting time on passenger loss is often overlooked in the aforementioned studies, which motivates \citet{mao2020dispatch} to introduce the definition of impatient passengers. The cancellation cost caused by user-specific tolerance of waiting time is regarded as one of the components of the reward function. The proposed model shows its superiority in reducing the cancellation rate and total waiting time of impatient passengers for the taxi system by the Actor-Critic framework.

\subsection{Order Dispatching}

On the premise of ensuring available vehicles in various areas by vehicle re-positioning, the dispatching strategies to meet the large volume of orders in real-time are emphasized in a large number of studies. Traditional rule-based solutions for order dispatching require sophisticated hand-crafted parameter design but are only effective on simplified problem settings \citep{li2019efficient}, which motivates the utilization of Reinforcement Learning.

\citet{oda2018movi} is among the first to examine the framework of DQN for order dispatching in MOD systems. The dispatch center is regarded as the agent to minimize the passenger waiting time and idle cruising time and reduce the number of requests that are not responded to. However, all idle vehicles need to sequentially decide their destinations which will increase computation time and decrease the dispatching efficiency. Thus, the following studies to be discussed consider the agent as the driver/vehicle to construct a multi-agent-based RL framework for order dispatching.

\citet{zhou2019multi} illustrates that explicit cooperation among various drivers is helpless for order dispatching since each driver serves different orders with different starting times, duration, and destination grids. Thus, each driver/vehicle is regarded as an agent working independently in this proposed method to explore the environmental information of the current locations, including the number of idle vehicles and valid orders and the destinations. To maximize the accumulated driver income (ADI) and order response rate (ORR), Double-DQN is extended with Kullback-Leibler (KL) divergence optimization to select optimal orders for drivers. 

More studies \citep{xu2018large, li2019efficient, he2019spatio} hold a different opinion with \citet{zhou2019multi}, which demonstrate the necessity of coordination among drivers for order dispatching. In detail, \citet{li2019efficient} clarifies that active agents sharing orders in the same/nearby areas might select the same order according to their own policy, which may cause conflicts. To solve such an issue, Mean Field Reinforcement Learning \citep{yang2018mean} is adopted to evaluate the average response among agents for agents interactions where the average response is derived from the number of drivers arriving at the same neighborhood and available orders. \citet{he2019spatio} proposes a capsule-based Double-DQN for coordination policy learning where the capsule means a structured group of neurons \citep{sabour2017dynamic}. The capsule construction helps the agent to analyze spatial (e.g., geographical distributions of demands and supplies) and temporal (e.g., weather conditions over time) relations and further learn the final policy. In addition, \citet{xu2018large} formulates the action-value as a bipartite graph matching problem (i.e., the edge between one driver and one order is set as the action-value). The Kuhn-Munkres (KM) algorithm \citep{munkres1957algorithms} is employed for optimization to ensure that each order is assigned to at most one driver and avoid conflicts.

The mass deployment of MOD systems shows great success and high profits in megalopolis motivate the popularization of MOD systems in tier-three cities, which lack data for optimization and management. Therefore, on the ride order dispatching problem, \citet{wang2018deep} and \citet{tang2019deep} propose transfer learning methods to enable knowledge transfer from source cities with sufficient historical data to target cities with limited historical data. Since travel patterns of different cities often share common spatial and temporal characteristics, reusing previously trained DQN models learned from source cities to determine the optimal policies for target cities can be flexible and useful. Three transfer learning methods are tested in these two studies, i.e., fine-tune \citep{hinton2006reducing}, progressive network \citep{rusu2016progressive}, and correlated-feature progressive transfer \citep{wang2018deep}.

The aforementioned studies dealing with order dispatching and vehicle re-positioning independently may ignore the high correlations between them \citep{jin2019coride}. \citet{holler2019deep} and \citet{jin2019coride} explore these two tasks (order dispatching and vehicle re-positioning) simultaneously, which means actions of an agent (i.e., driver/vehicle) contain vehicle re-positioning without an order and orders serving. \citet{holler2019deep} aims to maximize the revenue of each driver independently from driver-perspective and maximize the combined revenue across all drivers from system-perspective by using different reward specifications and optimization algorithms (i.e., DQN and PPO). The optimization results show that the driver-perspective system is more competitive than the system-perspective approach. It is noteworthy that most multi-agent-based RL methods designed for MOD systems management regard each driver/vehicle as an agent, which results in high computational costs due to a large number of agents. Based on the framework of Hierarchical RL, \citet{jin2019coride} chooses the region as an agent where large districts are manager agents while small grids are worker agents to model the ride-hailing system. The goal of the manager agent is to maximize ADI and ORR based on observations and peer messages (i.e., features extracted from other manager agents). The worker manager generates actions (i.e., pick up orders or re-position) following the objective developed by its manager and own observations.

The pricing strategy is overlooked in a large number of papers, which is studied in \citet{chen2019inbede} by optimizing pricing and dispatching strategies jointly with TD-based RL. In an online ride-hailing system, the user decides whether to submit the order request after knowing the estimated price of the input trip (i.e., origin and destination) given by the system. The request will be sent to the order list if the passenger approves the price. Therefore, the action value of order dispatching depends on environmental states (e.g., locations of drivers and passengers) and pricing strategies. In addition, the total expected reward of the pricing strategy is composed of expected driver income before order completion and actual driver income, which means the optimal pricing strategy also relies on order dispatching. 

More recently, \citet{manchella2021flexpool} presents a novel and valuable direction for joint goods delivery and ride-sharing service with deep RL methods. Using the status of available vehicles and pick-up requests, the proposed model adopts Double-DQN to find optimal dispatching policies for the pooling of passengers and delivering of goods. The ride-sharing data collected from New York City taxi-cab and customer check-in traffic data from Google Maps give the opportunity for this work to verify that jointly serving passengers and goods can be cost-efficient and environmentally friendly.

\subsection{Vehicle Routing}

In ride-sharing systems, multiple orders and various passengers with similar itineraries can be handled simultaneously, which means that the policies for vehicle routing after order dispatching should be addressed and studied. The methods with computational complexity issues are hard to be applied in time-sensitive vehicle routing applications. RL has already shown strong capabilities in vehicle routing/navigation. Also, the training process of RL-based strategies can be conducted offline so that the route generation process can be handled handy and fast \citep{james2019online} in large transportation networks. Therefore, RL becomes an essential tool for vehicle routing in ride-sharing service systems.

\citet{james2019online} tries to serve more orders while minimizing the driving distances of all vehicles. Based on the formulation of green logistic systems \citep{james2017autonomous}, the Asynchronous Advantage Actor-Critic (A3C) method is adopted to train the route construction policy with a dispatching center working as the agent. To further explicitly study the cooperation or competition among vehicles or customers, \citet{zhang2020multi} regards each vehicle as an agent and designs a multi-agent attention RL-based model. The model consists of an encoder-decoder structure where the encoder module analyzes the relations among customers while the decoder module decides the choice of the next visited customer via reinforcing gradient estimator optimization. The optimization of vehicle routing independently neglects the correlations between order dispatching and vehicle routing, which motivates \citet{al2019deeppool} to focus on providing policies for two tasks simultaneously via Double-DQN. Each vehicle works as an agent to decide whether to serve existing or new users after observing and analyzing the predicted future demand and the time cost before vehicles become available. If a new user is chosen or the vehicle is empty, the agent determines the zone to arrive. This study shows the superiority of ride-sharing in reducing traffic congestion through experiments on the real-world dataset from New York City.

\section{Assistant and Autonomous Driving} \label{sec:av}

Ensuring safety is the most critical objective in transportation systems for both human-piloted driving and autonomous driving. Driver-assistance systems (DASs) and autonomous vehicles (AVs) are expected to enhance driving safety and also improve traffic efficiency \citep{pan2021integrated}. In this section, a widely studied DAS technology, adaptive cruise control (ACC), with the strategies of Reinforcement Learning, is introduced first. Then, two types of training methods for decision-making modeling based on RL (i.e., car-following modeling to decide the velocity/acceleration and lane-changing modeling for steering control \citep{ye2019automated}) are presented. A list of studies using RL for assistant/autonomous driving is provided in Table~\ref{table:av}.

\begin{table}[]
\caption{Summary of RL Applications in Assistant and Autonomous Driving}
\begin{threeparttable}
\setlength{\tabcolsep}{.01mm}{
\begin{tabular}{c|c|c|c|c|c}
\hline
\textbf{Reference} & \textbf{Application} & \textbf{Framework} & \textbf{Agent} & \textbf{\begin{tabular}[c]{@{}c@{}} Scenario/\\Data \end{tabular}} & \textbf{Simulator} \\ \hline
\citet{desjardins2011cooperative} & \begin{tabular}[c]{@{}c@{}} adaptive cruise \\ control \end{tabular} & DDPG & \begin{tabular}[c]{@{}c@{}}single-agent, \\ a vehicle \end{tabular} & \begin{tabular}[c]{@{}c@{}} synthetic \\ network \end{tabular} & \begin{tabular}[c]{@{}c@{}} personal \\ simulator \end{tabular} \\ \hline
\citet{li2017training} & \begin{tabular}[c]{@{}c@{}} adaptive cruise \\ control \end{tabular} & Q-learning & \begin{tabular}[c]{@{}c@{}}single-agent, \\ a vehicle \end{tabular} & \begin{tabular}[c]{@{}c@{}} synthetic \\ network \end{tabular} & \begin{tabular}[c]{@{}c@{}} personal \\ simulator \end{tabular} \\ \hline
\citet{li2019ecological} & \begin{tabular}[c]{@{}c@{}} adaptive cruise \\ control \end{tabular} & \begin{tabular}[c]{@{}c@{}}Deep \\ Actor-Critic \end{tabular} & \begin{tabular}[c]{@{}c@{}}single-agent, \\ a vehicle \end{tabular} & \begin{tabular}[c]{@{}c@{}} synthetic \\ network \end{tabular} & \begin{tabular}[c]{@{}c@{}} personal \\ simulator \end{tabular} \\ \hline
\citet{gao2019reinforcement} & \begin{tabular}[c]{@{}c@{}}adaptive cruise \\ control for buses \end{tabular} & \begin{tabular}[c]{@{}c@{}} Policy \\ Iteration \end{tabular} & \begin{tabular}[c]{@{}c@{}}single-agent, \\ the center \end{tabular} & \begin{tabular}[c]{@{}c@{}} Lincoln \\ Tunnel \\ Corridor \end{tabular} & Paramics \\ \hline
\citet{nascimento2021development} & \begin{tabular}[c]{@{}c@{}} drivers' comfort \\ modeling \end{tabular} & Double-DQN & \begin{tabular}[c]{@{}c@{}}single-agent, \\ a vehicle \end{tabular} & \begin{tabular}[c]{@{}c@{}} synthetic \\ network \end{tabular} & \begin{tabular}[c]{@{}c@{}} GTA V \\ simulator \tnote{1} \end{tabular} \\ \hline
\citet{zhu2018human} & \begin{tabular}[c]{@{}c@{}} acceleration \\ control \end{tabular} & DDPG & \begin{tabular}[c]{@{}c@{}}single-agent, \\ a vehicle \end{tabular} & \begin{tabular}[c]{@{}c@{}} synthetic \\ network \end{tabular} & \begin{tabular}[c]{@{}c@{}} personal \\ simulator \end{tabular} \\ \hline
\citet{zhou2019development} & \begin{tabular}[c]{@{}c@{}} acceleration \\ control \end{tabular} & DDPG & \begin{tabular}[c]{@{}c@{}}single-agent, \\ the center \end{tabular} & \begin{tabular}[c]{@{}c@{}} synthetic \\ network \end{tabular} & \begin{tabular}[c]{@{}c@{}} personal \\ simulator \end{tabular} \\ \hline
\citet{zhu2020safe} & \begin{tabular}[c]{@{}c@{}} velocity control \\ for electric vehicle\end{tabular} & DDPG & \begin{tabular}[c]{@{}c@{}}two agents, \\ following and \\ lead vehicle\end{tabular} & \begin{tabular}[c]{@{}c@{}} NGSIM \\ dataset \tnote{2} \end{tabular} & \begin{tabular}[c]{@{}c@{}} Next Generation \\ Simulation \tnote{2} \end{tabular} \\ \hline
\citet{wegener2021automated} & \begin{tabular}[c]{@{}c@{}} acceleration \\ control \end{tabular} & \begin{tabular}[c]{@{}c@{}} Twin-delayed \\ DDPG \end{tabular} & \begin{tabular}[c]{@{}c@{}}single-agent, \\ a vehicle \end{tabular} & \begin{tabular}[c]{@{}c@{}} NGSIM \\ dataset \end{tabular} & \begin{tabular}[c]{@{}c@{}}Intelligent \\ Driver Model \end{tabular} \\ \hline
\citet{liu2021reinforcementdriving} & lane keeping & DDPG & \begin{tabular}[c]{@{}c@{}}single-agent, \\ a vehicle \end{tabular} & \begin{tabular}[c]{@{}c@{}} real and \\ synthetic \\ scenarios \end{tabular}& \begin{tabular}[c]{@{}c@{}} simulator from \\ OpenAI Gym\end{tabular} \\ \hline
\citet{cao2020highway} & \begin{tabular}[c]{@{}c@{}}acceleration control \\ and lane changing \\ for highway existing\end{tabular} & \begin{tabular}[c]{@{}c@{}} Monte Carlo \\ Tree Search \end{tabular} & \begin{tabular}[c]{@{}c@{}}single-agent, \\ a vehicle \end{tabular} & \begin{tabular}[c]{@{}c@{}} synthetic \\ network \end{tabular} & \begin{tabular}[c]{@{}c@{}} personal \\ simulator \end{tabular} \\ \hline
\citet{ye2019automated} & \begin{tabular}[c]{@{}c@{}}acceleration control \\ and lane changing\end{tabular} & DDPG & \begin{tabular}[c]{@{}c@{}}single-agent, \\ a vehicle  \end{tabular} & \begin{tabular}[c]{@{}c@{}} synthetic \\ network \end{tabular} & VISSIM \\ \hline
\citet{guo2021hybrid} & \begin{tabular}[c]{@{}c@{}}acceleration control \\ and lane changing\end{tabular} & DDPG & \begin{tabular}[c]{@{}c@{}}single-agent, \\ a vehicle  \end{tabular} & \begin{tabular}[c]{@{}c@{}} synthetic \\ network \end{tabular} & SUMO \\ \hline
\citet{pan2021integrated} & \begin{tabular}[c]{@{}c@{}}ramp metering, \\ lane changing, \\ speed limit control\end{tabular} & \begin{tabular}[c]{@{}c@{}}Cross- \\ Entropy- \\ Method \end{tabular} & \begin{tabular}[c]{@{}c@{}}single-agent, \\ a vehicle \end{tabular} & \begin{tabular}[c]{@{}c@{}} synthetic \\ network \end{tabular} & \begin{tabular}[c]{@{}c@{}} personal \\ simulator \end{tabular} \\ \hline
\citet{wachi2019failure} & \begin{tabular}[c]{@{}c@{}} failure scenario \\ finding \end{tabular} & DDPG & \begin{tabular}[c]{@{}c@{}}multi-agent, \\ a vehicle \end{tabular} & \begin{tabular}[c]{@{}c@{}} synthetic \\ network \end{tabular} &
\begin{tabular}[c]{@{}c@{}}Microsoft AirSim, \\ \citep{shah2018airsim} \end{tabular} \\ \hline
\end{tabular}}
\begin{tablenotes}
\footnotesize
\item[1] https://github.com/aitorzip/DeepGTAV
\item[2] https://ops.fhwa.dot.gov/trafficanalysistools/ngsim.htm
\end{tablenotes}
\end{threeparttable}
\label{table:av}
\end{table}

\subsection{Adaptive Cruise Control}

The technologies of driver-assistance systems have been embedded into vehicles to improve the driving experience and reduce traffic accidents. Adaptive cruise control (ACC), as an essential function of the system, has the ability to adjust the speed and acceleration of the current vehicle and further maintain a safe distance from the vehicle in front of it. To reduce reliance on prior knowledge of disturbance measurements \citep{li2017training}, Reinforcement Learning becomes a valuable tool for ACC.

Two types of vehicles are investigated for adaptive cruise control with RL, i.e., the car and bus. As for the car, \citet{desjardins2011cooperative} and \citet{li2017training} collect the speed and acceleration of the current vehicle and the distance from the front vehicle as the state for adaptive cruise control policy optimization. \citet{desjardins2011cooperative} takes advantage of DDPG to determine the action (e.g., braking, accelerating). \citet{li2017training} utilizes Q-learning to select the specific values of permissive accelerations, which can be more feasible in practice. Moreover, \citet{li2019ecological} investigates driving safety and fuel consumption simultaneously by optimizing the velocity and the online gear shift jointly. The utilized deep Actor-Critic framework consists of two actor networks and a critic network. Two actor networks are used to generate the traction force for velocity tracking and provide the gear position for fuel economy, respectively. And the critic network evaluates the control performance for these two purposes.

As for the bus, \citet{gao2019reinforcement} proposes a cooperative ACC algorithm with a central controller for a fleet of autonomous buses on the exclusive bus lane (XBL). The policy iteration RL method is employed to approximate the value of the control gain introduced in the linear optimal control theory \citep{lewis2012optimal}. The experimental results show that the proposed method is able to increase the traffic throughput and save the travel time of buses.

More recently, \citet{nascimento2021development} reports that safe driving can be affected by the driver's comfort and feel, which can be adaptable for all types of vehicles. To investigate the interplay between the perceived sounds of a vehicle and the driver’s attention/enjoyment, a psychoacoustic (PA) metric \citep{pedersen2008many} is used as the reward function to measure the driver's feeling where lower PA values mean more comfort. The agent analyzes environmental sounds (e.g., pedestrians and traffic) and noises (e.g., sounds of bells and beeps) to decide the states of the window (no change, open, close), radio (no change, on, off), and speed (no change, accelerate, decelerate) with the optimization via Double-DQN. The proposed method has the ability to change the state of the vehicle to maintain the driver's concentration for driving safety.

\subsection{Velocity and Acceleration Control}

Velocity/acceleration control of the autonomous vehicle has the promise of improving traffic safety and increasing road capacity \citep{zhu2020safe}, which has been studied in numerous studies with Reinforcement Learning.

\citet{zhu2018human} introduces an autonomous driving model based on DDPG to reproduce the behaviors and trajectories of drivers. To determine the acceleration of the vehicle, the agent sets the reward function as minimizing the disparity of spacing and velocity between the simulated and observed data. Note that solely imitating human driving behaviors for autonomous vehicles may not reduce traffic accidents or increase road capacity due to the hardly optimal operation of human drivers \citep{zhu2020safe}. Thus, the following studies \citep{zhou2019development, zhu2020safe, wegener2021automated} directly optimize autonomous driving from interactions with the simulated environment (i.e., surrounding vehicles information, own driving information, and road networks) by adopting deep RL strategies. DDPG is adopted in \citet{zhou2019development, zhu2020safe} while Twin-delayed Deep Deterministic Policy Gradient (TD3) \citep{fujimoto2018addressing} is used by \citet{wegener2021automated}. \citet{zhou2019development} and \citet{wegener2021automated} focus on obtaining appropriate driving acceleration under different levels of traffic and lengths of the signal cycle at intersections. \citet{zhu2020safe} examines velocity control of autonomous driving under different road incidents/events, which improves safety, efficiency, and comfortableness, as shown by their experimental results.

\subsection{Steering Control and Lane Changing}

Keeping the vehicle within the lane and driving stably are essential for the safety of autonomous driving  \citep{liu2021reinforcementdriving}. Specifically, \cite{liu2021reinforcementdriving} collects the distances from the vehicle to the road lane borders from the GPS information as the state to decide the vehicle’s steering angle via the framework of DDPG. To accommodate the real-world scenario with information noise, a noise compensation approach is used.

The effectiveness of steering control and velocity control attracts the following works to combine the management of two tasks with RL. Many studies determine longitudinal and lateral positions simultaneously to achieve safer and more efficient autonomous driving. Initial works only depend on one optimizing strategy for two tasks \citep{ye2019automated, cao2020highway}. In detail, in order to increase the success rate of exiting from highways in heavy dynamic traffic, \citet{cao2020highway} optimizes longitudinal acceleration and the policy of lane changing by Monte Carlo Tree Search \citep{browne2012survey} and regarding the distance to the exit ramp and the surrounding vehicles’ positions and speeds as the state.  \citet{ye2019automated} proposes a more general strategy to decide the longitudinal and lateral position jointly of the vehicle based on DDPG with the driving information of surrounding vehicles. The reward form is calculated by its distance from the preceding vehicle, its speed, and the speed difference to the preceding vehicle. The collision, uncomfortableness, and inefficient driving performances are also penalized in the reward. In addition, \citet{guo2021hybrid} also focuses on determining positions in both directions, which finds the optimal policies for the continuous longitudinal acceleration/deceleration and discrete lane changing via DDPG and DQN, respectively. The two optimizing strategies are able to interact with each other and reduce the error probability, which is more robust in unusual driving conditions (e.g., abrupt deceleration of the front vehicle). 

Furthermore, an integrated model is proposed to deal with more comprehensive tasks, i.e., ramp metering, variable speed limit, and lane changing control for both connected autonomous vehicles and regular human-piloted vehicles to minimize the total travel cost in \citet{pan2021integrated}. The proposed model is optimized by the gradient-free Cross-Entropy-Method-based algorithm \citep{szita2006learning}.

In addition, \citet{wachi2019failure} deals with the safety of autonomous driving in a novel manner by identifying failure scenarios of the vehicle. The environment consists of two types of vehicles, the player and multiple non-player characters (NPCs). And the aim is to train NPCs to make the player cause an accident or arrive at the destination late. When the player fails, NPCs get the adversarial reward based on their own contributions to the failure. Multi-agent DDPG algorithm \citep{lowe2017multi} is employed to train the agents to find the optimum driving directions and velocity. Their strategy provides a novel and effective direction to avoid catastrophic accidents for autonomous driving.

\section{Routing} \label{sec:routing}

In Section~\ref{sec:taxi}, RL-based vehicle routing in taxi, ride-sourcing, and ride-sharing systems have been reviewed. This section discusses RL-based routing in a more general context, where routing plays an important role in both human-driving and autonomous driving vehicles. Many previous studies on routing problems are based on parametric models with strong behavior assumptions \citep{mao2018reinforcement}. Instead, given its capability for optimal policy discovery without expert knowledge and its scalability for adapting the proposed methods to large-scale real-world networks, RL-based models have been used to find the shortest path and minimize total travel time. The introduced RL-based works for routing are summarized in Table~\ref{table:routing}.

\begin{table}[]
\caption{Summary of RL Applications in Routing}
\setlength{\tabcolsep}{.01mm}{
\begin{tabular}{c|c|c|c|c|c}
\hline
\textbf{Reference} & \textbf{Application} & \textbf{Framework} & \textbf{Agent} & \textbf{Scenario/Data} & \textbf{Simulator} \\ \hline
\citet{cao2017maximizing} & \begin{tabular}[c]{@{}c@{}} path \\ recommendation \end{tabular} & DQN & \begin{tabular}[c]{@{}c@{}}single-agent, \\ the driver\end{tabular} & \begin{tabular}[c]{@{}c@{}} networks of Munich, \\ Singapore, Beijing\end{tabular} & \begin{tabular}[c]{@{}c@{}} personal \\ simulator \end{tabular} \\ \hline
\citet{boutilier2018planning} & \begin{tabular}[c]{@{}c@{}} shortest path \\ routing \end{tabular} & DQN & \begin{tabular}[c]{@{}c@{}}single-agent, \\ the driver\end{tabular} & \begin{tabular}[c]{@{}c@{}}network in San \\ Francisco Bay Area\end{tabular} & \begin{tabular}[c]{@{}c@{}} personal \\ simulator \end{tabular} \\ \hline
\citet{chandak2020reinforcement} & \begin{tabular}[c]{@{}c@{}} shortest path \\ routing \end{tabular} & \begin{tabular}[c]{@{}c@{}} Policy Gradient \\ algorithm \end{tabular} & \begin{tabular}[c]{@{}c@{}}single-agent, \\ the driver\end{tabular} & \begin{tabular}[c]{@{}c@{}} network in San \\ Francisco Bay Area\end{tabular} & \begin{tabular}[c]{@{}c@{}} personal \\ simulator \end{tabular} \\ \hline
\citet{mao2018reinforcement} & \begin{tabular}[c]{@{}c@{}}routing for \\ travel time \\ minimization\end{tabular} & \begin{tabular}[c]{@{}c@{}}Neural fitted \\ Q-iteration, \\ Q-learning\end{tabular} & \begin{tabular}[c]{@{}c@{}}single-agent, \\ the driver\end{tabular} & \begin{tabular}[c]{@{}c@{}} Sioux Falls \\ network \end{tabular} & \begin{tabular}[c]{@{}c@{}} personal \\ simulator \end{tabular} \\ \hline
\citet{silva2019reinforcement} & \begin{tabular}[c]{@{}c@{}} vehicle routing with \\ time window \end{tabular} & Q-learning & \begin{tabular}[c]{@{}c@{}} multi-agent, \\ a vehicle \end{tabular} & synthetic data & \begin{tabular}[c]{@{}c@{}} personal \\ simulator \end{tabular} \\ \hline
\citet{zhang2020increasing} & \begin{tabular}[c]{@{}c@{}} GPS \\ correctness \end{tabular} & A3C & \begin{tabular}[c]{@{}c@{}}single-agent, \\ the controller\end{tabular} & \begin{tabular}[c]{@{}c@{}}GPS trip recorder \\ in Southeast \\ Michigan\end{tabular} & \begin{tabular}[c]{@{}c@{}} personal \\ simulator \end{tabular} \\ \hline
\citet{an2020space} & \begin{tabular}[c]{@{}c@{}}routing for \\ travel time \\ minimization\end{tabular} & DQN & \begin{tabular}[c]{@{}c@{}}single-agent, \\ the controller\end{tabular} & synthetic data & \begin{tabular}[c]{@{}c@{}} personal \\ simulator \end{tabular} \\ \hline
\citet{zhang2019reinforcement} & parking & DDPG & \begin{tabular}[c]{@{}c@{}} single-agent, \\ the controller \end{tabular} & synthetic data & \begin{tabular}[c]{@{}c@{}} personal \\ simulator \end{tabular} \\ \hline
\citet{wang2021online} & parking & Monte-Carlo & \begin{tabular}[c]{@{}c@{}} single-agent, \\ the controller \end{tabular} & synthetic data & \begin{tabular}[c]{@{}c@{}} personal \\ simulator \end{tabular} \\ \hline
\end{tabular}}
\label{table:routing}
\end{table}

The stochastic shortest path (SSP) problem studied in \citet{cao2017maximizing} solves the issue of low accuracy and high computational cost caused by the previous tail-based research \citep{lim2013practical}. The proposed model adopts Q-learning as the framework and designs an approximator (i.e., a two-layer Neural Network) to represent the value function which can be well adapted to large road networks. In practice, some travel paths are not always reachable due to road construction or other reasons, which motivates the exploration of the unavailability of actions. Thus, \citet{boutilier2018planning} investigates the Markov decision processes with stochastic action sets (SAS) and illustrates the effects on the shortest path sought problem via DQN with the consideration of the probability of the shortest path availability. The results indicate that the optimal policy with SAS has the ability to yield an expected travel time between the origin and destination within a target small range. Following the study of \citet{boutilier2018planning}, \citet{chandak2020reinforcement} further examines each node as the origin and learns the shortest path from each node. The proposed framework generalizes the policy gradient algorithm to estimate the optimal policy in a large-scale network.

RL methods for the SSP problem build the foundation for the real-time routing strategy, which needs to minimize the expected total travel time by accounting for real-time traffic conditions. Therefore, continuous variables describing real-time traffic congestion are used in \citet{mao2018reinforcement, an2020space} to look for the path that minimizes travel time or travel delay. Neural Fitted Q Iteration (FQI) \citep{ernst2005tree} is adopted in \citet{mao2018reinforcement} to accommodate the large state space (i.e., the constantly changing instantaneous travel cost) and produce a more refined representation of the Q-function for further routing policy optimization. Differently, \citet{an2020space} proposes a platoon strategy to assist routing rather than determine the next location with RL directly. The aim of the platoon strategy is to avoid conflict points in platoons, where DQN is utilized to determine the platoon size on the monitor link. Dijkstra algorithm and k-shortest path algorithm are designed for routing.

Moreover, routing for parking issues has been discussed in \citet{zhang2019reinforcement, wang2021online}. Specifically, \citet{zhang2019reinforcement} adopts DDPG for autonomous parking (i.e., determine the steering wheel angle) with the coordinates of the four corner points in the vehicle. \citet{wang2021online} proposes a Monte-Carlo-based optimization model on parking spot selections, which becomes a crucial problem in mega-cities for automated multistory parking facilities. In order to reduce customers’ waiting time, the agent is in charge of choosing the parking level for each vehicle on the elevator by analyzing the status of available parking spots and the current time. 

Another application of vehicle routing, i.e., determining a set of routes to make each customer can be served by one vehicle based on a single depot, is explored in \citet{silva2019reinforcement} with Q-learning. To minimize the number of vehicles and reduce travel distances, each vehicle is regarded as an agent to decide the locations and order of passengers to be served by acknowledging the information of all vehicles and customers.

The accuracy of Global Positioning System (GPS) localization is also critical in vehicle navigation/routing applications, which might be affected by environmental factors (e.g., weather and occlusion of buildings). Thus, \citet{zhang2020increasing} proposes to correct raw GPS observations (i.e., longitude and latitude coordinates) by the algorithm of Asynchronous Advantage Actor-Critic. The state is the observation history trajectory consisting of the last reported position and the most recent predicted positions within a certain period. The robot introduced in simultaneous localization and mapping \citep{bailey2006consistency} provides the ground truth information of localization and evaluates the proposed model.

\section{Public Transportation and Bike-sharing System} \label{sec:public}

The public transportation system (e.g., buses and trains) and bike-sharing system serve a large number of passengers on a daily basis and play a vital role \citep{li2021urban} in metropolitan areas. RL-based strategies have been examined for public transit and bike-sharing systems scheduling and management to improve efficiency and profitability, which are reviewed in this section. A summary of the papers to be discussed is provided in Table~\ref{table:public}.

\begin{table}[h]
\caption{Summary of RL Applications in Public Transportation and Bike-sharing System}
\setlength{\tabcolsep}{.01mm}{
\begin{tabular}{c|c|c|c|c|c}
\hline
\textbf{Reference} & \textbf{Application} & \textbf{Framework} & \textbf{Agent} & \textbf{Data} & \textbf{Simulator} \\ \hline
\citet{alesiani2018reinforcement} & bus holding & Double-DQN & \begin{tabular}[c]{@{}c@{}}multi-agent, \\ a bus\end{tabular} & \begin{tabular}[c]{@{}c@{}}a main bus line \\ in Singapore\end{tabular} & \begin{tabular}[c]{@{}c@{}} personal \\ simulator \end{tabular} \\ \hline
\citet{chen2016real} & bus holding & Q-learning & \begin{tabular}[c]{@{}c@{}}multi-agent, \\ a bus\end{tabular} & synthetic data & \begin{tabular}[c]{@{}c@{}} personal \\ simulator \end{tabular} \\ \hline
\citet{menda2018deep} & bus holding & PS-TRPO & \begin{tabular}[c]{@{}c@{}}multi-agent, \\ a bus\end{tabular} & synthetic data & \begin{tabular}[c]{@{}c@{}} personal \\ simulator \end{tabular} \\ \hline
\citet{wang2020dynamic} & bus holding & deep PPO & \begin{tabular}[c]{@{}c@{}}multi-agent, \\ a bus\end{tabular} & synthetic data & \begin{tabular}[c]{@{}c@{}} personal \\ simulator \end{tabular} \\ \hline
\citet{yin2014intelligent} & \begin{tabular}[c]{@{}c@{}}acceleration \\ control for \\ the subway\end{tabular} & Q-learning & \begin{tabular}[c]{@{}c@{}}single-agent, \\ a subway\end{tabular} & \begin{tabular}[c]{@{}c@{}}real data from \\ Beijing Subway \end{tabular} & \begin{tabular}[c]{@{}c@{}} personal \\ simulator \end{tabular} \\ \hline
\citet{yang2020deep} & \begin{tabular}[c]{@{}c@{}} voltage control for \\ urban railway \end{tabular} & DQN & \begin{tabular}[c]{@{}c@{}}single-agent, \\ the center \end{tabular} & \begin{tabular}[c]{@{}c@{}}real data from \\ Beijing Subway \end{tabular} & \begin{tabular}[c]{@{}c@{}} personal \\ simulator \end{tabular} \\ \hline
\citet{vsemrov2016reinforcement} & train scheduling & Q-learning & \begin{tabular}[c]{@{}c@{}}single-agent, \\ the center \end{tabular} & \begin{tabular}[c]{@{}c@{}}railway network \\ in Slovenia\end{tabular} & \begin{tabular}[c]{@{}c@{}} personal \\ simulator \end{tabular} \\ \hline
\citet{khadilkar2018scalable} & train scheduling & Q-learning & \begin{tabular}[c]{@{}c@{}}single-agent, \\ the center \end{tabular} & \begin{tabular}[c]{@{}c@{}}railway lines \\ from Indian \end{tabular} & \begin{tabular}[c]{@{}c@{}} personal \\ simulator \end{tabular} \\ \hline
\citet{ying2020actor} & subway scheduling & DDPG & \begin{tabular}[c]{@{}c@{}}single-agent, \\ the center \end{tabular} & \begin{tabular}[c]{@{}c@{}} London \\ Underground \end{tabular} & \begin{tabular}[c]{@{}c@{}} personal \\ simulator \end{tabular} \\ \hline
\citet{jiang2018reinforcement} & \begin{tabular}[c]{@{}c@{}} inflow control for \\ urban rail transit\end{tabular} & Q-learning & \begin{tabular}[c]{@{}c@{}}single-agent, \\ the center \end{tabular} & \begin{tabular}[c]{@{}c@{}} metro line \\ in Shanghai\end{tabular} & \begin{tabular}[c]{@{}c@{}} personal \\ simulator \end{tabular} \\ \hline
\citet{wei2020city} & \begin{tabular}[c]{@{}c@{}} next metro \\ line design \end{tabular} & \begin{tabular}[c]{@{}c@{}}Deep \\ Actor-Critic \end{tabular} & \begin{tabular}[c]{@{}c@{}}single-agent, \\ the center \end{tabular} & \begin{tabular}[c]{@{}c@{}}the current metro \\ network in Xi’an \end{tabular} & \begin{tabular}[c]{@{}c@{}} personal \\ simulator \end{tabular} \\ \hline
\citet{li2018dynamic} & \begin{tabular}[c]{@{}c@{}} bike re-position\\ for bike-sharing \\ system\end{tabular} & DQN & \begin{tabular}[c]{@{}c@{}}multi-agent, \\ a trike\end{tabular} & \begin{tabular}[c]{@{}c@{}} Citi Bike data \\ from New York \end{tabular} & \begin{tabular}[c]{@{}c@{}} personal \\ simulator \end{tabular} \\ \hline
\citet{pan2019deep} & \begin{tabular}[c]{@{}c@{}} price management \\ for bike-sharing \\ system\end{tabular} & \begin{tabular}[c]{@{}c@{}}DDPG, \\ Hierarchical RL \end{tabular} & \begin{tabular}[c]{@{}c@{}}multi-agent, \\ a user \end{tabular} & \begin{tabular}[c]{@{}c@{}} Mobike dataset \\ from Shanghai \end{tabular} & \begin{tabular}[c]{@{}c@{}} Mobike’s \\ original \\ system \end{tabular} \\ \hline
\end{tabular}}
\label{table:public}
\end{table}

\subsection{Bus Holding}

Bus holding, a strategy that delays buses at control points \citep{dai2019predictive}, has received substantial attention for many decades in order to reduce the probability of bus delay, decrease the waiting/travel time of passengers, and thus improve the efficiency of the bus system \citep{berrebi2018comparing}. A large number of strategies mainly consider local information with a pre-specified headway/schedule. However, the global coordination of the whole bus fleet and the long-term effect are often overlooked \citep{wang2020dynamic}, which can be potentially addressed by RL-based methods.

Owing to the mutual influence among buses, existing studies \citep{chen2016real, alesiani2018reinforcement, menda2018deep, wang2020dynamic} adopt the multi-agent RL framework by regarding each bus as an agent to analyze the input state (e.g., treating departure time, arrival time, and target headway time of the bus). \citet{alesiani2018reinforcement} uses Double-DQN to decide the duration of bus holding. Due to constraints from real-world driving conditions, a bus holding time less than 30 seconds is not practical \citep{chen2016real}. Thus, \citet{chen2016real} introduces a Q-learning-based framework for the distributed coordinated bus holding control to choose the holding time as some multiple of the holding time unit (e.g., 30 seconds). Furthermore, considering Neural Networks are more capable of handling non-linear relations, \citet{menda2018deep} utilizes a deep-based RL model, PS-TRPO \citep{gupta2017cooperative}, to optimize policies for the bus holding time selection. Though these methods adopt multi-agent frameworks to deal with holding time for multiple buses simultaneously, less attention has been paid to agents' cooperation. More recently, \citet{wang2020dynamic} proposes a global joint action tracker into the PPO framework to incorporate global coordination for dynamic bus holding control. The action tracker network is used to adopt the global information of buses and passengers to further track the policies of each agent (i.e., a bus). Thus, the state evaluation of each agent’s policy is based on the local environment and other agents' decisions.

\subsection{Urban Rail Transit System Management}

Adopting the mechanism of Reinforcement Learning, multiple research topics have been investigated for the operation of the urban rail transit system (e.g., train and subway), such as energy management, vehicle re-scheduling, and passenger flow control, which will be introduced in this subsection. 

\textbf{Energy management:} In order to minimize the energy consumption of subway operation, \citet{yin2014intelligent} defines the subway control model as a Markov decision process where the state includes the current vehicle position, the speed, and the reserved trip time. Each subway vehicle works as an agent that decides the variation of acceleration in the time interval via Q-learning, which is hard to cooperate with other subways to acknowledge the time-vary traffic. More practically, \citet{yang2020deep} uses the super-capacitor energy management system (SCESS) as the central agent for energy-saving and voltage stabilization of multiple subways. The states of the subways near the SCESS and the rectifier current/voltage of the substation where the SCESS is installed account for the state in the implementation of RL. And the agent decides on the combination of charging and discharging voltage threshold to increase the energy-saving rate and voltage stabilization rate in each time step.

\textbf{Scheduling:} Scheduling is one of the core issues for urban rail transit systems, e.g., in order to reduce the travel/waiting time and the operating cost \citep{zhao2021integrated}. Train scheduling of the railway is explored in \citet{vsemrov2016reinforcement} with Q-learning to show its superiority in reducing the total delay. Considering the information in relation to the locations of trains, the infrastructure availability of block sections, and the time, the proposed model decides the actions for each signaling element, i.e., setting it to red (stop) or green (go) color, to decide which trains can move on to the next section. However, their study only deals with the single-track railway, which cannot be directly adapted to multi-track railway systems (e.g., the trains operating on multiple tracks can be merged into one track which may cause disruption). Train scheduling on multi-track is taken into consideration by the study of \citet{khadilkar2018scalable}, where the directions of trains' motion are analyzed for further decision-making with Q-learning. More recently, to reduce passenger waiting times and subway operating costs, the scheduling of subway is explored in \citet{ying2020actor} by taking the number of passengers into account for decision-making via DDPG. 

\textbf{Passenger flow control:} To decrease the waiting time of passengers and reduce accidents caused by crowds in railway stations, the control of passenger inflow for railway has been investigated in \citet{jiang2018reinforcement}. The environmental state includes information of real-time passenger demand, the arrival/departure time, the available capacity of trains, and the platform capacity of stations. And Q-learning is adopted to set the rate of inflow volume for each station. The experimental results show that inflow control with RL can reduce the number of passengers being stranded and relieve passenger congestion at certain stations.

\textbf{Network expansion:} The design or the expansion of a railway transit network is another primary concern in public rail/transit systems \citep{laporte2010game}. Most existing strategies dealing with network expansion are often based on conventional mathematical programming approaches, which are heavily dependent on expert guidance and behavior assumptions \citep{wei2020city}. Instead of the usage of domain knowledge and behavior assumptions, the Actor-Critic framework with single-agent is adopted in \citet{wei2020city} to select the locations of expanded stations in the city metro network. Specifically, the actor network is an Encoder-Decoder Neural Network coupling with an attention layer to parameterize the station selection policy for expanding the metro line, while the critic network consists of three convolutional layers and two fully connected layers to estimate the expected cumulative reward of the next metro line.

\subsection{Bike-sharing System}

Bike-sharing systems, including dock and dock-less systems, are widely deployed in urban and rural areas to ease the first/last-mile problems and reduce the usage of private vehicles. \citet{li2018dynamic} and \citet{pan2019deep} aim to balance the supply and demand of these two systems, respectively. In order to minimize the customer loss of the system with dock, \citet{li2018dynamic} proposes a multi-agent DQN-based bike re-positioning method. Each trike (i.e., the tool for moving bikes) is regarded as the agent that chooses the location of the station and the number of picking up or unloading bikes after observing the system status (i.e., bike and dock availability at each station), its own status (i.e., the available location for bikes), and the status of other trikes. \citet{pan2019deep} focuses on pricing management to incentive users for the dock-less bike-sharing system. Building upon DDPG and Hierarchical RL, the proposed pricing algorithm suggests the user return the bike to neighboring regions by offering a price incentive under a default budget.

\section{Electric Vehicle: Energy Management, Charging, and Ride Service} \label{sec:ee}

To mitigate the crisis of resource scarcity and climate change, electrification has been the trend of the automotive industry to achieve the merits of high performance and long-term economy \citep{wu2020battery}. Reinforcement learning methods have been adopted for electric vehicle (EV) control and management in recent years, especially for ground electric vehicles. This section mainly introduces the RL applications on two major ground vehicles, hybrid electric vehicles (HEVs) (including plug-in HEVs) and pure-electric vehicles. The mentioned works in this study are summarized in Table~\ref{table:ev}.

\begin{table}[]
\caption{Summary of RL Applications in Electric Vehicle}
\begin{threeparttable}
\setlength{\tabcolsep}{.8mm}{
\begin{tabular}{c|c|c|c|c|c}
\hline
\textbf{Reference} & \textbf{Application} & \textbf{Framework} & \textbf{Agent} & \textbf{Data} & \textbf{Simulator} \\ \hline
\citet{liu2015reinforcement} & \begin{tabular}[c]{@{}c@{}}fuel and electricity \\ sources control\end{tabular} & Q-learning & \begin{tabular}[c]{@{}c@{}}single-agent, \\ a vehicle \end{tabular} & synthetic data & MotoTune \tnote{1} \\ \hline
\citet{qi2016data} & \begin{tabular}[c]{@{}c@{}}fuel and electricity \\ sources control\end{tabular} & Q-learning & \begin{tabular}[c]{@{}c@{}}single-agent, \\ a vehicle \end{tabular} & \begin{tabular}[c]{@{}c@{}}inductive loops detector \\ data archived in \\ the California \\ Freeway PEMS \tnote{2} \end{tabular} & \begin{tabular}[c]{@{}c@{}} Motor Vehicle \\ Emission \\ Simulator \tnote{3} \end{tabular} \\ \hline
\citet{liu2017reinforcement} & \begin{tabular}[c]{@{}c@{}}fuel and electricity \\ sources control\end{tabular} & Q-learning & \begin{tabular}[c]{@{}c@{}}single-agent, \\ a vehicle \end{tabular} & synthetic data & \begin{tabular}[c]{@{}c@{}} personal \\ simulator \end{tabular} \\ \hline
\citet{qi2019deep} & \begin{tabular}[c]{@{}c@{}}fuel and electricity \\ sources control\end{tabular} & \begin{tabular}[c]{@{}c@{}} DQN \\ Dueling-DQN \end{tabular} & \begin{tabular}[c]{@{}c@{}}single-agent, \\ a vehicle \end{tabular} & \begin{tabular}[c]{@{}c@{}}inductive loops detector \\ data archived in \\ the California \\ Freeway PEMS \tnote{2} \end{tabular} & \begin{tabular}[c]{@{}c@{}} personal \\ simulator \end{tabular} \\ \hline
\citet{wu2019deep} & \begin{tabular}[c]{@{}c@{}}fuel and electricity \\ sources control\end{tabular} & DDPG & \begin{tabular}[c]{@{}c@{}}single-agent, \\ a vehicle \end{tabular} & synthetic data & Paramics \\ \hline
\citet{lian2020rule} & \begin{tabular}[c]{@{}c@{}}fuel and electricity \\ sources control\end{tabular} & DDPG & \begin{tabular}[c]{@{}c@{}}single-agent, \\ a vehicle \end{tabular} & synthetic data & \begin{tabular}[c]{@{}c@{}} personal \\ simulator \end{tabular} \\ \hline
\citet{wan2018model} & \begin{tabular}[c]{@{}c@{}}EV charging/ \\ discharging \\ scheduling\end{tabular} & DQN & \begin{tabular}[c]{@{}c@{}}single-agent, \\ a vehicle \end{tabular} & \begin{tabular}[c]{@{}c@{}}real scenario from \\ the California ISO\end{tabular} & \begin{tabular}[c]{@{}c@{}} personal \\ simulator \end{tabular} \\ \hline
\citet{zhang2020effective} & \begin{tabular}[c]{@{}c@{}}EV charging/ \\ discharging \\ scheduling\end{tabular} & DQN & \begin{tabular}[c]{@{}c@{}}single-agent, \\ a vehicle  \end{tabular} & \begin{tabular}[c]{@{}c@{}}real data from EV \\ charging stations data \\ in Beijing \end{tabular} & \begin{tabular}[c]{@{}c@{}} personal \\ simulator \end{tabular} \\ \hline
\citet{luo2020rebalancing} & EV re-positioning & PPO & \begin{tabular}[c]{@{}c@{}}multi-agent, \\ a hexagonal \\ grid\end{tabular} & \begin{tabular}[c]{@{}c@{}}real EV sharing data \\ in Shanghai\end{tabular} & \begin{tabular}[c]{@{}c@{}} personal \\ simulator \end{tabular} \\ \hline
\citet{shi2019operating} & \begin{tabular}[c]{@{}c@{}} EV dispatching \\ and charging\\ management \end{tabular} & DQN & \begin{tabular}[c]{@{}c@{}}multi-agent, \\ a vehicle\end{tabular} & synthetic data & \begin{tabular}[c]{@{}c@{}} personal \\ simulator \end{tabular} \\ \hline
\citet{tang2020online} & \begin{tabular}[c]{@{}c@{}}EV taxi-customer \\ assignments, vehicle \\ dispatching \\ and charging\end{tabular} & Deep RL & \begin{tabular}[c]{@{}c@{}}single-agent, \\ a central \\ controller\end{tabular} & \begin{tabular}[c]{@{}c@{}}real data from \\ Tongzhou and Beijing \end{tabular} & \begin{tabular}[c]{@{}c@{}} personal \\ simulator \end{tabular} \\ \hline
\citet{zhang2020route} & \begin{tabular}[c]{@{}c@{}} EV route planning \\ and energy \\ management \end{tabular} & \begin{tabular}[c]{@{}c@{}} Actor-Critic, \\ Q-learning \end{tabular} & \begin{tabular}[c]{@{}c@{}} single-agent, \\ the controller \end{tabular} & synthetic data &  ADVISOR \tnote{4} \\ \hline
\citet{lin2021deep} & \begin{tabular}[c]{@{}c@{}} vehicle routing \\ for Electric Vehicles \end{tabular} & REINFORCE & \begin{tabular}[c]{@{}c@{}} single-agent, \\ the controller \end{tabular} & synthetic data & \begin{tabular}[c]{@{}c@{}} personal \\ simulator \end{tabular} \\ \hline
\end{tabular}}
\begin{tablenotes}
\footnotesize
\item[1] http://mcs.woodward.com/support/wiki/index.php?title=MotoTune
\item[2] http://pems.dot.ca.gov
\item[3] https://www.epa.gov/moves
\item[4] http://bigladdersoftware.com/advisor/docs/advisor$\_$doc.html
\end{tablenotes}
\end{threeparttable}
\label{table:ev}
\end{table}

\subsection{Hybrid Electric Vehicle}

A hybrid electric vehicle usually combines a conventional powertrain (e.g., gasoline) with an electric engine. Most existing studies dealing with energy management of hybrid EVs follow pre-defined rules, which heavily rely on the accurate prediction of future traffic conditions and are not straightforward for applications under time-sensitive driving conditions \citep{qi2019deep}. RL strategies have been effective tools to avoid the need for precise forecasts.

The studies start to regard the energy management center as the agent for engine power control via Q-learning in \citet{liu2015reinforcement, qi2016data, liu2017reinforcement} with different state settings. In detail, \citet{liu2015reinforcement} explores the knowledge of environmental features, the battery state-of-charge (SOC), and the rotational speed of the generator (i.e., engine speed) to determine fuel consumption. More related characteristics are analyzed in \citet{qi2016data}, i.e., the vehicle velocity, road grade, percentage of remaining time to destination, SOC, and available charging gain of the selected charging station. The internal combustion engine (ICE) power supply level (discrete form) is chosen to further control the proportions of electricity and fuel to use. The predicted future velocity profile and the information of SOC are utilized in \citet{liu2017reinforcement} as the state to select the throttle engine power and further determine the power distribution of the electrical energy source and conventional powertrain source. And the velocity profile is obtained by two novel velocity predictors (i.e., Nearest Neighbor Velocity Predictor and Fuzzy Encoding Velocity Predictor).

A number of deep RL studies have shown their capability to handle non-linear and complicated relations among vehicles and the traffic environment for traffic control, which motivates the utilization of deep learning in energy management. \citet{qi2019deep, wu2019deep, lian2020rule} adopt complex and powerful deep RL methods to control electricity and conventional powertrain energy split for HEVS. To reduce fuel consumption, \citet{qi2019deep} uses DQN and Dueling-DQN to select an optimal fuel/electricity split's level (i.e., 24 power level outputs of the engine) with the knowledge of the power demand at the wheel, the battery pack’s state-of-charge, and the distance to the destination. This study optimizes the agents based on a single driving cycle that might not be able to deal with different driving cycles (DCs) or the entire driving profile of a vehicle \citep{wu2019deep}. Therefore, \citet{wu2019deep} adopts the framework of DDPG to model the energy split management for multiple driving cycles. Given the control variables (e.g., rational speed of engine/motor) as the current state of the environment, the actor network represented by the structured control net (SCN) \citep{srouji2018structured} produces an action while the critic network consisting of several fully connected layers estimates the action-value function. Moreover, considering that human expertise can provide optimal training samples or preferences for the learning agent to guide exploration in the training process, \citet{lian2020rule} proposes a rule-interposing DDPG model to deal with the time-consuming problem caused by deep RL strategies. The added expert knowledge includes the optimal brake specific fuel consumption (BSFC) curve of the HEV engine and the battery characteristics, which helps set the control variable of RL models. The aim of the controller is to optimize the engine power increment or decrement (e.g., remain unchanged, increase one kilowatt, decrease one kilowatt).

Different from the aforementioned studies focusing on energy management and splitting independently, \citet{lin2021deep} adopts the Actor-Critic framework and Q-learning for route planning with power management of plug-in HEVs to minimize energy consumption. The inner loop is in charge of managing power by controlling the desired output torque from the engine, the gear shift command, and the direction by analyzing the state (i.e., vehicle status and geographic information). Meanwhile, the outer loop decides the changes in road slope and vehicle speed, which can affect energy utilization. The overall reward is designed to minimize fuel consumption and battery recuperation instead of only considering the shortest distance between the origin and the destination.

\subsection{Pure-Electric Vehicle} 

The usage of pure-electric vehicles is rapidly growing, while the driving range and insufficient charging stations of EVs are two adverse factors on the widespread adoption of Pure-Electric vehicles \citep{he2018optimal}. In order to solve such issues, recently, \citet{wan2018model, zhang2020effective} design the DQN-based frameworks for EV charging/discharging scheduling with different optimization aims. To improve user benefit, \citet{wan2018model} proposes a representation network to extract discriminative features from the battery state-of-charge (SOC) and the future price trends predicted by Long Short-Term Memory (LSTM). The Q-network is utilized to approximate the optimal action-value function and then make the decision for the amount of energy that the EV battery will be charged or discharged. Moreover, to minimize the total charging time of EVs and reduce the distance between the origin and charging station, the EVs charging schedule system in \citet{zhang2020effective} analyzes the features from the available charging piles and the EVs electricity consumption (predicted by distance traveled with linear regression) to obtain Q-value for selecting a charging station for the vehicle. 

Pure-Electric vehicles have also been introduced to provide ride-sourcing services with the fast improvement of battery technologies and the rapid growth of recharging facilities \citep{kim2015factors, ke2019modelling}. As presented in Section~\ref{sec:taxi}, a number of RL-based methods have been put into use for dispatching and routing gasoline vehicles, which can also be adapted for ride-sourcing management of EVs. Different from conventional gasoline vehicles, EV re-position, dispatching, and routing often more explicitly take into account the recharging issues or electricity consumption issues of EVs.

Specifically, unbalanced/skewed distributions of EV fleets motivate \citet{luo2020rebalancing} to propose a multi-agent RL model for EV re-positioning in order to improve demand rate and net revenue. The designed actor-critic-PPO model consists of two connected policy networks, one used for choosing the grid and another adopting the output from the first network for further selecting the station in the chosen grid with the agent (i.e., each hexagon grid of the urban area in concern). The proposed model can deal with the non-stationarity in action spaces caused by the station extension or closure. The reward function is regularized to deal with the non-stationary reward problem.

Vehicle dispatching for an electric vehicle is studied in \citet{shi2019operating, tang2020online}. \citet{shi2019operating} designs a DQN-based algorithm to dispatch the electric vehicle for ride-hailing services in terms of reducing EV operational costs and customer waiting time. The proposed framework consists of two components: the decentralized learning process to approximate the state-value function with the knowledge of vehicles and dispatching tasks; the centralized decision-making process formulates the state-value function for EV fleets and maximizes the action-value by a linear assignment problem to find the optimal dispatching policy. \citet{tang2020online} designs a two-step framework, advisor-student RL, to dispatch vehicles and arrange charging activities. In the advisor network, the control center assigns the status of vehicles (i.e., to be charged or to accept the order) to minimize the system cost (i.e., customer waiting cost, customer abandon penalty, vehicle travel cost, and vehicle charging cost) through the optimization by DQN. The student network decides the vehicle-customer pair and vehicle-charging-station pair via assignment problem optimization. 

With the different aim of \citet{shi2019operating, tang2020online}, \citet{lin2021deep} focuses on reducing total distances of electric vehicles by solving routing problems (i.e., choosing the geographical coordinate of the next location) with the REINFORCE algorithm \citep{williams1992simple}.

\section{Future Directions and Conclusion} \label{sec:challenge}

In the past decade, we have seen a growing number of studies (a growing trend) that develop/adapt Reinforcement Learning methods for applications in the transportation sector. However, the development and utilization of advanced RL strategies for a more efficient and sustainable transportation system are still at an early stage. This section will discuss several aspects that require substantial efforts in terms of developing RL methods for real-world transportation applications, i.e., scalability, practicality, transferability, and fairness. \\

\noindent \textbf{Scalability:} 
\begin{itemize}
\item Existing RL-based studies for transportation applications are capable of dealing with a single subject and/or one aspect of the system (e.g., speed limit control for a target part of the freeway \citep{zhu2014accounting}). However, the need for computing resources and computing time can be extremely high when adapting these methods to multiple-object large-scale environments, especially where there are complex interactions among objects or sub-systems within the system (e.g., a city often is served with thousands of intersections). Developing competent models with a cooperative and/or competitive multi-agent RL-based framework to deal with multi-object large-scale transportation systems is crucial. For instance, handling a single bus or train in bus bunching control and urban rail transit system management will be more feasible given the current development of RL methods, while optimizing the whole system with a large number of objects (or agents) will be much more challenging. Developing a scalable model with the ability to adopt and analyze large-scale spatial-temporal features and jointly optimize the actions of multi-object requires substantial novel efforts and innovations. Hierarchical RL can be a promising concept for handling such large-scale problems with a centralized manager for overall control and optimization and multiple decentralized workers for implementations at the local level. 
\end{itemize}

\noindent \textbf{Practicality:} 
\begin{itemize}
\item The design of the environment and reward function is imperative for RL-based methods. Many methods have shown the superiority of the simulations with simulated observations and rewards. Only several works take advantage of real-world platforms (e.g., \citet{zhou2019multi} uses the platform provided by Didi Chuxing for order dispatching) for evaluation. Nevertheless, it may leave a certain (and unknown) gap with actual feedback. It is essential to conduct and evaluate the proposed methods in real-world environments to produce real rewards for policy optimization. For instance, order dispatching for MOD systems can be efficiently tested on real-world platforms such as Uber and Didi Chuxing so that the actual values of order response rate, driver income, and waiting/travel time can be obtained. Also, the utilization of a digital twin framework to mimic the real network as a virtual network can be helpful in obtaining real feedback. This often requires coordinated and cooperative efforts from academia, industry, and government. 
\item Existing studies are able to avoid soft constraints effectively by introducing the penalty to reward functions (e.g., \citet{tang2020online} introduces a customer abandon penalty to reduce the possibility of order cancellation). Unfortunately, the hard constraints of the environment are tough to be eliminated through the reward penalty, which should be investigated in future studies. The design of environments and actions with limitations may be a feasible approach. For instance, the number of moving bikes in the bike-sharing system cannot exceed the capacity of the trike (the tool for moving bikes), which can be achieved by designing the range of the action vector.
\item The evaluation of RL methods relying on ideal simulated environments has achieved satisfactory progress (e.g., bus holding without considering the sluggish of passengers \citep{alesiani2018reinforcement}). In practice, uncertainties, disruptions, and accidents often occur for road traffic, rail traffic, and air traffic. External factors which may influence the transportation system and network traffic should be analyzed or predicted (e.g., accurate weather forecasting can effectively help aircraft scheduling), and then incorporated in RL methods for better optimization and control schemes. For example, disturbances and information noises in the real world (e.g., pedestrians breaking out suddenly on the road and unexpected changes of weather) should be reproduced to reflect the real-world environment while testing. 
\item Some information such as travel demand, traffic flow, vehicle speed, trip distance, trip time might be simulated or estimated for further decision-making with RL methods (e.g., Citi Bike demand data from New York is collected in \citet{li2018dynamic} for bike re-positioning). However, precise information of some specific characteristics in the environment may not be readily available or hard to be obtained. For example, some existing research for energy management of the electric vehicles may require precise information regarding the drivers' behaviors, which might not be available at the time of decision-making. Therefore, some estimation or expectations might have to be assumed or further methods without such information request have to be developed (\citet{qi2019deep, wu2019deep}). 
\item Some existing methods using discrete formulations for environmental features (e.g., the level of traffic congestion) and actions (e.g., slow down or speed up in adaptive cruise control \citep{desjardins2011cooperative}), achieved satisfactory performance based on private and public simulators. This is likely not universal and might not be sufficient in many real-world occasions. The extension of such methods to other applications might not be feasible or might result in low quality solutions. It is necessary to propose methods that are able to deal with the continuity and granularity of actions in transportation and optimize the choice of continuity and granularity since different scenarios require continuous or discrete actions with different (optimal) granularity. For example, the acceleration and steering control for autonomous driving requires extremely precise decisions since a slight adjustment in steering may cause a large change in the direction of a vehicle in the case of high-speed driving. On the contrary, it might be less meaningful to have a holding time for buses of ten seconds (while ten seconds might be too long for autonomous driving applications).
\item The optimization of the decision-making for one type of action may help improve systems performance in many studies (e.g., the inflow control for urban rail transit in \citet{jiang2018reinforcement}). Nevertheless, the isolated design of different types of actions also eliminates the practicality of RL to solve more complex transportation problems with substantial endogeneity or correlations among actions. Studies dealing with only one or two specific aspects of autonomous driving (e.g., lane changing, motion control, and collision avoidance) is still not ready for practical applications. More comprehensive consideration of multi-type actions simultaneously can be critical and essential in solving more complicated transportation problems in future research (e.g., to ensure safe, reliable, and efficient autonomous driving, the velocity, acceleration, angle change, route, and passengers' preference might have to be examined in an integrated manner).
\end{itemize}

\noindent \textbf{Transferability:} 
\begin{itemize}
\item Studies targeting on existing road networks and public transit routes/stations have shown great success in numerous aspects, such as train scheduling \citep{khadilkar2018scalable} and routing \citep{mao2018reinforcement}. Due to urban expansion, new transportation facilities have to be designed and arranged in existing and new regions, which receives less attention in the literature. The construction of new facilities requires sufficient expert knowledge due to the scarcity of historical data for policy optimization in RL. The utilization of transfer learning \citep{pan2009survey} and Meta-based RL \citep{finn2018meta} (i.e., the combination of Meta-Learning and Reinforcement Learning) are potentially effective tools for organizing new tasks or applications that lack sufficient training data. These strategies are able to transfer/adapt the trained RL-based model parameters/policies learned from the regions that already have related facilities to the new model for the new regions to help the model training.
\end{itemize}

\noindent \textbf{Fairness:} 
\begin{itemize}
\item Previous works aiming at improving the efficiency of systems have made admirable progress in diverse transportation applications (e.g., minimizing overall wait time at intersections in adaptive traffic signal control \citep{chen2020toward}). However, the fairness issue of transportation systems has not been considered much, and the right of each entity to be selected (e.g., the signals of certain intersections are red for a long time) is ignored in the development of RL methods. All targets or entities (e.g., intersections or vehicles) need to be fairly treated in the formulation of RL. To better address fairness issues in transportation, exploring the combination of survey data (stated preference) and other multi-source data is necessary. How to incorporate such a combination of data into RL method development is a direction worth further examination. Therefore, combinational weighted rewarding optimization problems with multiple objectives might have to be considered and addressed in transportation applications to achieve both efficiency and fairness. The combinational weighted rewards are hard to be designed (e.g., the safety, efficiency, and comfort in autonomous driving are hard to be evaluated simultaneously), which can be solved by other algorithms. For instance, Inverse Reinforcement learning may be an effective solution to learning the reward function based on the agent's decisions and then finding the optimal policy.
\end{itemize} 

Reinforcement Learning and smart transportation are research topics that aroused substantial interest in recent years, where we see a large number of novel developments on strategies and techniques. It is also noted that applications of Reinforcement Learning in some sub-domains are limited, e.g., air traffic control and the aviation sector. For these application sub-domains, examining relevant and useful features is necessary to support relevant policy optimization. 

In summary, this paper first uses the bibliometric analysis to identify the development of RL methods for transportation applications in recent years and then provides a review of the most relevant works covering a wide range of topics. This review provides readers with an understanding of RL-based method developments and applications in smart transportation and can serve as a reference point for researchers interested in interdisciplinary Reinforcement Learning research in transportation and computer science.

\section*{Acknowledgments} The authors would like to thank all anonymous referees for their thoughtful and constructive comments, which have helped to improve this paper substantially. Dr Liu would like to acknowledge the support from The Hong Kong Polytechnic University (P0039246, P0040900, P0041316).

\bibliographystyle{apalike}

\bibliography{cas-refs}
\end{document}